\documentclass[final,5p,times,twocolumn]{elsarticle}
\usepackage[hyphens]{url}  
\usepackage{url}
\usepackage{mathtools}
\usepackage{multirow}
\usepackage{makecell}
\usepackage[linesnumbered,ruled,vlined]{algorithm2e}
\usepackage{graphicx}
\usepackage{footmisc}
\usepackage{subfigure}
\usepackage{float}
\usepackage{longtable} 
\usepackage{supertabular}
\usepackage{booktabs}
\usepackage{setspace}
\usepackage{xspace}
\newcommand{\mytablefontsize}{8pt}
\newcommand{\mytablebaselineskip}{0.6}

\newcommand{\kestabcolsep}{4pt}

\usepackage{xcolor}

\newcommand{\eg}{\emph{e.g.,}\xspace}
\newcommand{\ie}{\emph{i.e.,}\xspace}

\newcommand{\comment}[1]{}

\definecolor{hhcolor}{rgb}{0.9,0.55,0}

\definecolor{clcolor}{rgb}{0.5,0.5,0.8}

\definecolor{yccolor}{rgb}{0.9,0.1,0.1}

\definecolor{newcolor}{rgb}{0.9,0.4,0}

\begin{document}
\let\WriteBookmarks\relax
\def\floatpagepagefraction{1}
\def\textpagefraction{.001}

\title{Improving the Performance of Stochastic Local Search for Maximum Vertex Weight Clique Problem Using Programming by Optimization}

\author[1,2]{Yi Chu}
\author[3]{Chuan Luo} 
\author[4]{Holger H. Hoos}
\author[3]{Qingwei Lin}
\author[1]{Haihang You\corref{cor1}}

\address[1]{State Key Laboratory of Computer Architecture, Institute of Computing Technology, Chinese Academy of Sciences, Beijing 100190, China}
\address[2]{University of Chinese Academy of Sciences, Beijing 100049, China}
\address[3]{Microsoft Research, China}
\address[4]{Leiden Institute of Advanced Computer Science, Leiden University, Leiden, The Netherlands}
\cortext[cor1]{Corresponding author}

\begin{abstract}
The maximum vertex weight clique problem (MVWCP) is an important generalization of the maximum clique problem (MCP) that has a wide range of real-world applications.
In situations where rigorous guarantees regarding the optimality of solutions are not required, MVWCP is usually solved using stochastic local search (SLS) algorithms, which also define the state of the art for solving this problem. However, there is no single SLS algorithm which gives the best performance across all classes of MVWCP instances, and it is challenging to effectively identify the most suitable algorithm for each class of MVWCP instances.
In this work, we follow the paradigm of Programming by Optimization (PbO) to develop a new, flexible and highly parametric SLS framework for solving MVWCP, combining, for the first time, a broad range of effective heuristic mechanisms. 
By automatically configuring this PbO-MWC framework, we achieve substantial advances in the state-of-the-art in solving MVWCP over a broad range of prominent benchmarks, including two derived from real-world applications in transplantation medicine (kidney exchange) and assessment of research excellence.
\end{abstract}

\begin{keyword}
Maximum Clique Problem \sep Stochastic Local Search \sep Programming by Optimization
\end{keyword}

\maketitle

\section{Introduction}
\label{intro}
Given an undirected graph $G$, a clique is a subset of vertices of $C \subseteq G$ whose induced subgraph is complete.
The maximum clique problem (MCP) is to find a clique $C$ of maximum size $|C|$ in a given graph. MCP is one of the most widely known combinatorial optimization problems; it was one of the first problems proven to be NP-hard, with an NP-complete decision variant \cite{Karp1972Reducibility}. 
The maximum vertex weight clique problem (MVWCP) is an important generalization of the MCP, where each vertex is associated with a positive number representing its weight, and the objective is to find a clique with maximum weight, \ie{} a clique $C$ with maximum total weight over the vertices contained in $C$.
The MVWCP has a wide range of practical applications, including computer vision, pattern recognition, robotics \cite{Ballard1982Computer}, broadband network design \cite{Park1996An} and wireless telecommunication \cite{Balasundaram2006Graph}. 

In light of the importance of the MCP and MVWCP in theory and practice, considerable effort has been expended to develop effective algorithms for these problems. In practice, there are two popular categories of algorithms for solving the MCP and MVWCP: complete algorithms and incomplete algorithms. Complete algorithms are usually based on an approach known as \emph{branch and bound} (see, \eg{} \cite{ostergaard2001new,yamaguchi2008new,fang2016exact,jiang2017exact,jiang2018two}). These algorithms use sophisticated techniques to determine tight upper bounds and branching strategies, in order to reduce the search space and accelerate the search process.
In contrast, incomplete algorithms, which are mostly based on some form of local search, cannot prove the optimality of candidate solutions.
However, the best incomplete algorithms are usually able to find high-quality solutions even for large and challenging instances within reasonable time (see, \eg{} \cite{battiti2001reactive,grosso2004combining,pullan2006dynamic,pullan2008approximating,pullan2011cooperating,wu2012multi,WangCY2016,CaiL2016,zhou2017push,fan2017restart}). 

Among these incomplete algorithms, MN/TS \cite{wu2012multi} shows an advantage on the well-known BHOSLIB benchmark, where instances are generated in the phase-transition area according to Model RB \cite{XuEtAl05} and known to be difficult in theory and practice \cite{XuEtAl07}, while the strong configuration checking (SCC) strategy underlying the LSCC algorithm \cite{WangCY2016} exhibits stronger performance on the prominent DIMACS benchmark \cite{dimacs26}, which contains instances from applications in various areas. 
Overall, there is no single algorithm that performs best on all types of MVWCP instances, and selecting the most appropriate algorithm for a given set of instances is challenging. 


Designing an effective local search algorithm for MVWCP involves a large number of design choices, such as 1) how to construct the initial solution, 2) which intensification and diversification strategies to use, 3) how to strike a good balance between intensification and diversification, and 4) when to restart the local search process.
As noted earlier, to deal effectively with various types of MVWCP instances, different search strategies appear to be required, yet, current state-of-the-art MVWCP solvers are based on rather restrictive choices. 
In addition, for a given graph, it is quite challenging to identify the most effective combination of algorithmic strategies.

Recently, a novel algorithm design paradigm dubbed \emph{programming by optimization (PbO)} \cite{hoos2012programming} has been proposed to encourage algorithm developers to embrace and exploit rich design spaces incorporating a broad range of algorithmic techniques, to expose choices that may affect performance as parameters, and to use general-purpose automated configuration techniques to instantiate those choices such that performance for specific classes or sets of problems instances is optimized. Through the use of PbO, major improvements have been achieved in the state of the art for solving a broad range of NP-hard problems, including propositional satisfiability \cite{KhuEtAl16}, mixed integer programming \cite{hutter2010automated} and minimum vertex cover \cite{MVC2019luo}. 

In this work, we present what we believe to be the first application of the PbO paradigm to the MVWCP and achieve major improvements in the state of the art in solving this prominent and important NP-hard combinatorial optimization problem.
Our main contributions can be summarized as follows:

\begin{itemize}

\item We propose a new PbO-based local search framework for MVWCP dubbed \emph{PbO-MWC}, which is highly configurable and incorporates a broad range of of effective techniques. 

\item We conduct extensive experiments to compare PbO-MWC against five state-of-the-art solvers on four benchmarks, including two benchmarks that have been widely used in the literature (BHOSLIB and DIMAC), and two benchmarks derived from practical applications (kidney exchange and research excellence assessment). Our empirical results indicate that PbO-MWC outperforms its competitors on all four benchmarks. PbO-MWC significantly improves over the performance of state-of-the-art solvers for solving MVWCP on three benchmarks (BHSOLIB, kidney exchange and research excellence assessment).
Notably, usng PbO-MWC, we were able to achieve an improvement in the best known solution for a challenging graph known as `MANN\textunderscore a81'.

\item 
Based on our extensive empirical analysis, we provide insights into the efficacy of different  local search strategies and mechanisms.
\end{itemize}

The remainder of this paper is structured as follows. In Section \ref{sec:preliminaries}, we provide necessary definitions, notations and an introduction to multi-neighborhood search. In Section \ref{sec:relatedwork}, we give an overview of related work. Then, in Section \ref{sec:slsformwcp}, we present our PbO-MWC solver framework and provide detailed descriptions of all its core components. In Section \ref{sec:experimental_evaluations}, we conduct extensive experiments to show the effectiveness of PbO-MWC and analyze the experimental results. Finally, we conclude this paper and give future work in Section \ref{sec:conclusions}.
 
\section{Preliminaries}
\label{sec:preliminaries}
Given an undirected graph ${G = (V, E)}$, where $V = \{v_1,
v_2, \cdots, v_n\}$ is the set of vertices and $E = \{e_1, e_2, \cdots, e_m\}\subseteq V\times V$ is the set of edges, a clique $C$ is a subset of $V$, such that each pair of vertices in $C$ is connected by an edge. Given an undirected graph $G$, the objective in the maximum clique problem (MCP) is to find a clique $C \subseteq V$ with a maximum number of vertices. Given an undirected vertex-weighted graph $G=(V,E,w)$, where $(V,E)$ is an unweighted graph and $w$ a weight function that assigns a positive weight $w(v)$ to each vertex $v \in V$, the weight of a clique $C$ is defined as $w(C) := \Sigma _{v\in C} w(v)$. 
In the maximum vertex weight clique problem (MVWCP), the objective is to find a clique $C \subseteq V$ with maximum weight $w(C)$ in a given vertex-weighted graph $G$. 

As usual, edges $e \in E$ in a given undirected graph $G$ are represented as pairs of vertices $e(u,v)$, where $u,v \in C$ are called the end points of $e = (u,v)$.
Finally, for a vertex $v \in V$, we use ${N(v) := \{u\in V \mid (u,v)\in E\}}$ to denote the set of neighbours of $v$, and $d(v) := |N(v)|$ to denote the degree of $v$.

{
\SetAlFnt{\small}
\begin{algorithm} [t]
    {\caption{SLS for MVWCP}\label{alg:slsformwcp}}
    \KwIn{graph {\em G}=({\em V}, {\em E}, {\em W}), the cutoff time\;}
    \KwOut{$C^*$\;}
      initialize $C^*:=\emptyset, C:=\emptyset $\; 
      \While{no termination criteria are met}{
		
		\While{$V_{add}(C) \neq \emptyset$}{
		    v := select a vertex from $V_{add}(C)$\;
		    C := C $\cup$ \{v\}\;
		}
		\If{$W(C)$\textgreater $W(C^*)$}{
				$C^*:=C$\; 
		}
		\While{(no termination criteria are met) {\rm \textbf{and}} (no restarting criteria are met)}{
			{\em C := a neighboring clique of C}\;
			\If{$W(C)$\textgreater $W(C^*)$}{
				$C^*:=C$\; 
		    }
		}
	}
    \Return ${C^*}$\; 
\end{algorithm}
}

Stochastic local search (SLS) algorithms for the MVWCP usually build and subsequently modify cliques until some termination criterion is met, and then return the clique with the maximum weight encountered during the search process. 
Three kinds of search steps are performed: {\em add}, {\em swap} and {\em drop}. 
To formally define these, we introduce three sets of vertices for a given clique $C$:




\begin{itemize}
\item $V_{add}(C) = \{u \not\in C 
\mid \forall v \in C: u \in N(v)\}$, 
\ie{}, the set of vertices that, when added to $C$, result in a larger clique;

\item $V_{swap}(C) = \{u \not\in C \mid \exists v \in C: (u,v) \notin E \wedge  \forall w \in C\setminus \{v\}\}: (u,w) \in E\}$, 
\ie{}, the set of vertices that can replace some vertex in $C$ resulting in a new clique of the same size; 

\item $V_{drop}(C) = \{v \mid v \in C\}$, \ie{}, the set of all vertices that can be dropped from the given clique, resulting in a new, smaller clique.

\end{itemize}

An \emph{add} step moves from a clique $C$ to a clique $C \cup V_{add}(C)$; 
a \emph{swap} step moves from a clique $C$ to a clique $C \cup V_{swap}(C)$; 
and a \emph{drop} step moves from a clique $C$ to a clique $C \cup V_{drop}(C)$.
Cliques $C$ and $C'$ are called neighbours if $C'$ can be obtained from $C$ by a single \emph{add}, \emph{swap} or \emph{drop} step.

Formally, we define a general framework of SLS algorithms for the MVWCP as shown in Algorithm \ref{alg:slsformwcp}.
Starting from an empty set, we construct an initial clique {\em C} by iteratively choosing a vertex from $V_{add}(C)$ and add it into $C$, until $V_{add}(C) = \emptyset$ (Line 3-5); the choosing approach is based on different strategies. Then, we iteratively move from one clique to one of its neighbours until the termination criteria are met or some restarting criteria are met (e.g. a fixed number of steps have been iterated) (Line 8-11).







High-performance SLS algorithms usually balance between two types of strategies:  intensification and diversification strategies. In the case of MVWCP, intensification strategies aim to greedily improve the weight of the clique, \eg{} by iteratively moving from current clique to the neighbouring clique with the largest weight.
Diversification strategies are used to prevent or overcome search stagnation, usually by moving to a different part of the search space with little or no regard to solution quality (here: weight of the current clique).

Let {\em C} be a clique, $W(C)$ denotes the weight of clique {\em C}. We use {\em Ascore}({\em v},{\em C}) to denote the increment of $W(C)$ after adding {\em v} into {\em C}, i.e., {\em Ascore}{({\em v},{\em C})} = $W(C\cup \{v\})-W(C)$. We use {\em Sscore(u,v,C)} to represent the increment of $W(C)$ after both adding {\em u} into {\em C} and removing {\em v} from {\em C}, i.e., {\em Sscore}$(u,v,C)=W(C\backslash \{v\} \cup \{u\})-W(C)$. We use {\em Dscore}$(v,C)$ to express the increment of $W(C)$ after removing {\em v} from {\em C}, i.e., {\em Dscore}$(v,C)=W(C\backslash \{v\})-W(C)$.

Besides, given a graph $G=(V,E,W)$ and a clique $C$, a vertex $v\in V$ has two possible states: inside {\em C}, or outside {\em C}. We define the number of steps that has occurred since {\em v} last changed its states as the $age$ of {\em v}, denoted as $age(v)$.

\section {Related Work}
\label{sec:relatedwork}


In this section, we give an overview of related work, including the existing local search algorithms for MCP and MVWCP, programming by optimization and automated algorithm configuration, which form the basis of PbO-MWC.

\subsection{Existing Local Search Algorithms for MCP (MVWCP)}




Notable progress on local search algorithms for MCP and MVWCP has been made in recent years. In this subsection, we briefly review the most representative and the state-of-the-art local search algorithms. 
Reactive local search (RLS) \cite{battiti2001reactive}, dynamic local search (DLS-MC) \cite{pullan2006dynamic} and cooperating local search (CLS) \cite{pullan2011cooperating} are designed for MCP. RLS combines local-neighborhood-search with prohibition based diversification techniques. Starting from an empty clique, RLS explores the search space by two moves: {\em Add} and {\em Drop}. As soon as a vertex is added or dropped, it is put into the tabu list and remains prohibited for the next {\em T} iterations. The prohibition {\em T} is adjusted through feedback from the previous history of the search. RLS performs better than its predecessors on DIMACS benchmark. DLS-MC alternates between clique expansion phase and plateau search phase. The expansion phase selects a vertex from $V_{add}(C)$ to add it to the current clique. The plateau search phase selects a vertex pair from ${V_{swap}(C)}$ to perform the {\em Swap} move. The selection of vertices is based on vertex penalties that are dynamically adjusted during the search process. DLS-MC  shows excellent performance on DIMACS benchmark. CLS also alternates between clique expansion phase and plateau search phase. CLS integrates four low level heuristics which are effective for different instance types. The low level heuristics differ primarily in their vertex selection methods and also the perturbation mechanisms used to overcome search stagnation. CLS improves the state-of-the-art performance for MCP on BHOSLIB benchmark and achieves the performance that are comparable to state-of-the-art algorithms on DIMACS.

Multi-neighborhood tabu search (MN/TS) \cite{wu2012multi} is designed for MCP and MVWCP. Local search with SCC (LSCC) \cite{WangCY2016} and Restart and random walk in local search (RRWL) \cite{fan2017restart} are both developed on the basis of MN/TS and designed for solving MVWCP. These algorithms alternate between clique construction phase and and local search phase. In the local search phase, algorithms are based on a combined neighborhood induced by {\em Add}, {\em Drop} and {\em Swap} moves. The general prohibition rule in these algorithms is only prohibit removed vertices to move back to the clique {\em C} during the prohibition period, vertices that in {\em C} can be removed without restriction. In the local search phase, MN/TS adopts tabu strategy.
LSCC proposes a new prohibition mechanism named Strong Configuration Checking (SCC) that are based on Configuration Checking (CC). SCC mechanism is more restrictive than CC, a prohibited vertex will be lifted prohibition after adding one of its neighboring vertices to clique by {\em Add} move. LSCC shows obvious advantage on DIMACS benchmark when comparing with predecessors. RRWL \cite{fan2017restart} proposes a revisiting based restart strategy and adopts random walk strategy. In the local search phase, RRWL uses the SCC strategy. RRWL utilizes a hash table to record cliques that have been visited and detects revisiting after the first step which increases the clique weight. In order to increase diversity, RRWL adopts random walk strategy when choosing a vertex to perform {\em Drop} move.
RRWL finds a new best-known solution (111,341) on `MANN\textunderscore a81' instance from DIMACS benchmark. 

These algorithms do not contain various techniques and no single algorithm can perform well on all types of benchmarks. In addition, the performance of the above algorithms for solving MVWCP on many hard instances from BHOSLIB benchmark and DIMACS benchmark still exists the enhanced space. They also do not perform well on some benchmarks transformed from real-world problems.


\subsection{Programming by Optimization}
\label{subsec:programming_by_optimization}
PbO approach encourages algorithm developers to greatly expand the design space of algorithms by integrating more algorithmic technologies \cite{hoos2012programming}. Algorithm development following PbO approach usually involves exposing all design options as configuration parameters and searching for design alternatives to key components. 
The traditional algorithm configuration method is to test relatively few configurations through some experiments. With the progress of optimization and machine learning, solving the algorithm configuration problem as an optimization problem is a trend of algorithm design \cite{ansotegui2009gender,hutter2009paramils,HutterHL2011}. Up to now, PbO-based approaches have shown effectiveness in many problems, including Boolean satisfiability \cite{KhuEtAl16,hutter2017configurable}, mixed integer programming \cite{hutter2010automated,xu2011hydra}, AI planning \cite{vallati2013automatic} and Minimum Vertex Cover \cite{MVC2019luo}. 

\subsection{Automated Algorithm Configuration}
\label{subsec:automated_a_c}
The ability of complex heuristic algorithms to solve challenging combinatorial problem instances often critically depends on the use of suitable parameter settings \cite{hutter2009paramils}.
Since it may be difficult to seek for performance optimization values for these parameters, in recent years, some work has focused on the automated process of determining optimization parameter configurations, such as ParamILS \cite{hutter2009paramils}, GGA \cite{ansotegui2009gender}, F-RACE \cite{birattari2010f}, SMAC \cite{HutterHL2011} and irace \cite{L2016The}. In these automated algorithm configurations, SMAC is a sequential model-based algorithm configurator, that supports conditional parameters. Since SMAC is one of the best-performing algorithm configuration procedures, we utilized it to configure our PbO-MWC framework and the parametric competitors.

\section{Parametric Stochastic Local Search for MVWCP}
\label{sec:slsformwcp}

In order to explore the performance of configurations in configuration space
and avoid premature selection of strategies for different benchmarks, we design a stochastic local search framework named PbO-MWC for solving MVWCP, which exposes various strategies as parameters to the configuration procedure for selection. In the following, we first introduce the top-level design of PbO-MWC, then describe its key components, configuration space and default configuration.

\subsection{The PbO-MWC Framework}
\label{subsec:pbo-mwc-f}

The top level design of PbO-MWC can be described as follows: PbO-MWC iteratively executes two components until the termination criteria are met: initially, PbO-MWC generates an initialized clique; then PbO-MWC iteratively moves the current clique by local search step.

PbO-MWC consists of two components: construction and search. In the construction component, the initial clique is generated and regarded as the starting point of the search component. In the search component, PbO-MWC iteratively performs local search to move the clique to another.
The pseudo-code of PbO-MWC is outlined in Algorithm \ref{alg:framework}. There is an outer loop (lines 2-18 in algorithm \ref{alg:framework}) and an inner loop (lines 6-18 in algorithm \ref{alg:framework}). This framework determines whether the algorithm jumps from the inner loop to the outer loop through a parameter called {\em perform\textunderscore restart}, thereby restarting the local search process by reconstructing the initial clique. If the weight of current clique do not increase after performing {\em intensification\textunderscore process()} and {\em perform\textunderscore restart}= {\em True}, the algorithm restarts the search process with a probability of {\em restart\textunderscore prob} (lines 13-16 in algorithm \ref{alg:framework}).

\subsection{The Construction Component}
\label{subsec:construction_c}
To make construction effective, PbO-MWC adopts three simple effective construction approaches, resulting in three instantiations of the construction component {\em init\textunderscore construction} (Line 3 in algorithm \ref{alg:framework}):

i) Randomized approach: starting with an empty vertex set {\em C}, repeats randomly adding a vertex  $v\in V_{add}(C)$ into {\em C} until $V_{add}(C)=\emptyset$.

ii) Weight-based approach: randomly selecting a vertex {\em v} as the initialized clique {\em C}, repeats adding a vertex $v\in V_{add}(C)$ with the largest weight into {\em C} until $V_{add}(C)=\emptyset$.

iii) Degree-based approach: randomly selecting a vertex {\em v} as the initialized clique {\em C}, repeats adding a vertex $v\in V_{add}(C)$ with the largest degree into {\em C} until $V_{add}(C)=\emptyset$.

\subsection{The Random Walk Component}
\label{subsec:random_walk_c}
PbO-MWC utilizes a parameter called {\em perform\textunderscore randomwalk} to determine whether to attempt a random walk search or not, if {\em perform\textunderscore randomwalk}={\em true}, then PbO-MWC performs random walk search with a probability of {\em randomwalk\textunderscore prob}, where {\em randomwalk\textunderscore prob} is a parameter. This procedure is outlined in Algorithm \ref{alg:randomwalk}.

{
\SetAlFnt{\small}
\begin{algorithm} [t]
    {\caption{The PbO-MWC Framework}\label{alg:framework}}
    \KwIn{graph {\em G}=({\em V}, {\em E}, {\em W}), the cutoff time\;}
    \KwOut{$C^*$\;}
      initialize $C^* := \emptyset $\; 
      \While{no terminating criteria are met}{
		{\em C:=init}\textunderscore {\em construction()}\;
		\If{$W(C)$\textgreater $W(C^*)$}{
				$C^*:=C$\; 
		}
		\While{no terminating criteria are met}{
			\If{perform\textunderscore randomwalk}{
				\If{with probability randomwalk\textunderscore prob}{
					{\em C}:={\em random\textunderscore walk\textunderscore process(C)}\;	
					\textbf{continue}\;
				}
			}
			${C^{'}}$:={\em C}\;
			{\em C}:={\em intensification\textunderscore process(C)}\;
			\If{${W(C)\leq W(C^{'})}$}{
				\If{perform\textunderscore restart}{
					\If{with probability restart\textunderscore prob}{
						\textbf{break}\;
					}
				}
			}
			
			
			\ElseIf{$W(C)$\textgreater $W(C^*)$}{
				$C^*:=C$\; 
			}
		}
	}
       \Return ${C^*}$\; 
\end{algorithm}
}
{
\SetAlFnt{\small}
\begin{algorithm} [t]
    {\caption{The {\em random\textunderscore walk\textunderscore process} Procedure}\label{alg:randomwalk}}
    \KwIn{{\em C}\;}
    \KwOut{{\em C}\;}
      {\em prob}:=a random integer between 0 and 99\;
      \If{prob \textless 33 and $V_{add}(C) \neq \emptyset$}{
      	{\em v}:=a vertex randomly selected from $V_{add}(C)$\;
		$C:=C\cup \{v\}$\;
      } 
      \ElseIf{prob\textless 67 and $V_{swap}(C) \neq \emptyset$}{
      	$\left \langle u,v \right \rangle$:=a vertex pair randomly selected from $V_{swap}(C)$\;
		$C:=C\cup \{u\}\backslash\{v\}$\;
      }
      \Else{
      	{\em v}:=a vertex randomly selected from $V_{drop}(C)$\;
		$C:=C\backslash\{v\}$\;
      }
      \Return ${C}$\; 
\end{algorithm}
}

{
\SetAlFnt{\small}
\begin{algorithm} [t]
    {\caption{The {\em intensification\textunderscore process} Procedure}\label{alg:intensification}}
    \KwIn{{\em C}\;}
    \KwOut{{\em C}\;}
      {\em v} := a vertex with the largest {\em Ascore} in $V_{add}(C)$ and {\em v} is not forbidden by the prohibition mechanism, breaking ties by a {\em breakingTiesRule}\;
      \If{perform\textunderscore BMS}{
      	$\left \langle u,u' \right \rangle$ := a vertex pair with the largest {\em Sscore} in $V_{swap}(C)$ with BMS strategy and {\em u} is not forbidden by the prohibition mechanism, breaking ties by a {\em breakingTiesRule}\;
      }
      \Else{
      	$\left \langle u,u' \right \rangle$ := a vertex pair with the largest {\em Sscore} in $V_{swap}(C)$ and {\em u} is not forbidden by the prohibition mechanism, breaking ties by a {\em breakingTiesRule}\;
      }
      \If{$v \neq null$}{
      	\If{$\left \langle u,u' \right \rangle$=$\left \langle null,null \right \rangle$ or Ascore\textgreater Sscore}{
			$C:=C\cup \{v\}$\;
		}
		\Else{
			$C:=C\cup \{u\}\backslash\{u'\}$\;
		}
      } 
      \Else{
      	{\em x}:=a vertex with the largest {\em Dscore} in $V_{drop}(C)$\;
		\If{$\left \langle u,u' \right \rangle$=$\left \langle null,null \right \rangle$ or Dscore\textgreater Sscore}{
			$x$:=a vertex selected by a {\em selectDropVertexRule} from $V_{drop}(C)$\;
			$C:=C\backslash\{x\}$\;
		}
		\Else{
			$C:=C\cup \{u\}\backslash\{u'\}$\;
		}
      }
      \Return ${C}$\; 
\end{algorithm}
}

\begin{table*} [htp]
\begin{center}
\caption{The configuration space of PbO-MWC.}\label{tab:space}

\fontsize{\mytablefontsize}{\mytablebaselineskip}\selectfont\setlength{\tabcolsep}{\kestabcolsep}
\begin{spacing}{1}
\begin{tabular}{l  l  l  l  l}
\hline
 
Parameters & Depended Conditions & Parameter Type & Value Domain & Default Value \\ 
\hline

{\em perform\textunderscore BMS} & - & Categorical & \{{\em True},{\em False}\} & {\em True} \\
{\em bms\textunderscore num} & {\em perform\textunderscore BMS} = 1 & Integer & [1,100] & 50 \\
{\em breaking\textunderscore ties} & - & Categorical & \{0,1\} & 0 \\
{\em init\textunderscore construction} & - & Categorical & \{0,1,2\} & 0 \\
{\em drop\textunderscore vertex} & - & Categorical &  \{0,1,2\} & 0 \\
{\em randomdrop\textunderscore prob} & {\em drop\textunderscore vertex} = 1 & Categorical & \{0.1,0.2,...,0.9\} & 0.2 \\
{\em perform\textunderscore restart} & - & Boolean-valued & \{{\em True},{\em False}\} & {\em False} \\
{\em restart\textunderscore prob} & {\em perform\textunderscore restart} = 1 & Real & [0.0000001,0.0001] & 0.000001 \\
{\em perform\textunderscore randomwalk} & - & Boolean-valued & \{{\em True},{\em False}\} & {\em True} \\
{\em randomwalk\textunderscore prob} & {\em perform\textunderscore randomwalk} = 1 & Real & [0.00001,0.1] & 0.0001 \\	 
{\em tabu\textunderscore type} & - & Categorical & \{0,1,2\} & 1 \\
{\em tabu\textunderscore tenure} & {\em tabu\textunderscore type} = 1,2 & Integer & [1,100] & 7 \\
\hline

\end{tabular}
\end{spacing}

\end{center}
\end{table*}

\begin{table*} [htp]
\setlength{\abovecaptionskip}{-0.5cm}

\begin{center}
\caption{The default configuration of PbO-MWC}\label{tab:default}

\fontsize{\mytablefontsize}{\mytablebaselineskip}\selectfont\setlength{\tabcolsep}{\kestabcolsep}
\begin{spacing}{1}
\begin{tabular}{l l }

\hline
 
Instantiation & Default Configuration \\ 
\hline

\multirow{2}*{Default} & {\em perform\textunderscore BMS}={\em True}, {\em bms\textunderscore num}=50, {\em breaking\textunderscore ties}=0, {\em init\textunderscore construction}=0, {\em drop\textunderscore vertex}=0, \\ 
& {\em perform\textunderscore restart}={\em False}, {\em perform\textunderscore randomwalk}={\em True}, {\em randomwalk\textunderscore prob}=1.0E-4, {\em tabu\textunderscore type}=1, {\em tabu\textunderscore tenure}=7  \\

\hline

\end{tabular}
\end{spacing}

\end{center}
\end{table*}

\subsection {The Intensification Component}
\label{subsec:intensification_c}

The Intensification procedure is outlined in algorithm \ref{alg:intensification}. The {\em intensification\textunderscore process} moves the current clique to a neighboring clique with the maximum weight with prohibition mechanism. It first selects a vertex {\em v }$\in V_{add}(C)$ with the largest {\em Ascore} and {\em v} is not forbidden (Line 1), and selects a vertex pair $\left \langle u,u' \right \rangle$ $\in V_{swap}(C)$ with the largest {\em Sscore} and {\em u} is not forbidden (Line 2-5). If an {\em Add} move is possible, it selects the move with the largest weight (Line 6-10). On the contrary, it selects a vertex {\em x} $\in V_{drop}(C)$ with the largest {\em Dscore}, it picks the {\em Drop} move if there is no {\em Swap} move or {\em Dscore}\textgreater {\em Sscore} (Line 13-15), otherwise it picks the {\em Swap} move (Line 16-17).

In intensification search, local search algorithms correspond to efforts of revisiting promising regions of the search space \cite{HoosS2004}. In this process, algorithms may easily encounter the cycling phenomenon, i.e., returning to a candidate solution that has been visited recently. The cycling problem is an inherent problem of local search as the method does not allow our algorithms to memorize all previously visited candidate solutions. To deal with this severe issue, two fundamental prohibition mechanisms have been proposed to combine with local search
: tabu mechanism \cite{glover1989tabupart1} and configuration checking (CC) \cite{caiSS2011MVC}. Tabu mechanism was proposed by Glover \cite{glover1989tabupart1}, and the tabu mechanism forbids reversing the recent changes, where the ``strength" of prohibition is controlled by a parameter called {\em tabu\textunderscore tenure(tt)}. Configuration Checking (CC) was proposed by Cai \cite{caiSS2011MVC}, CC forbids a vertex to be added back into the candidate set until its circumstance information (also called configuration) has been changed. This paper adopts two prohibition mechanisms: one is based on the tabu mechanism and the other is derived from CC. A brief description is as follows:

The MN/TS algorithm proposes a prohibition rule: a vertex that leaves the current clique {\em C} (by a {\em Swap} or {\em Drop} move) is forbidden to move back to {\em C} for the next {\em tt} iterations. A vertex in the clique {\em C} is free to be removed from {\em C} without restriction \cite{wu2012multi}. 

For the {\em Swap} move, a vertex pair $\left \langle u,v \right \rangle\in V_{swap}(C)$, where {\em v} is removed from {\em C}, {\em v} is prohibited to be moved back to {\em C} for the next {\em tt\textunderscore swap} iterations. 

{\em tt\textunderscore  swap=random}$(|V_{swap}(C)|)+T_{1}$.

For the {\em Drop} move, a vertex $u\in V_{drop}(C)$, where {\em u} is removed from {\em C}, {\em u} is prohibited to be moved back to {\em C} for the next {\em tt\textunderscore drop} iterations.

{\em tt\textunderscore drop}=$T_{1}$.

$T_{1}$ is set to 7 as in \cite{wu2012multi}.

The LSCC algorithm proposes a prohibition rule called strong configuration checking (SCC) for MVWCP \cite{WangCY2016}. The SCC strategy is implemented with a Boolean array named {\em confChange}, where {\em confChange}[v] = 1 means {\em v} is allowed to be added to the current clique {\em C} and {\em confChange}[v] = 0 means {\em v} is forbidden to be added to {\em C}.

(1) Initially, for each vertex {\em v}, {\em confChange}[v] = 1.

(2) When {\em v} is added into {\em C}, {\em confChange}[u] = 1 for all $u\in N(v)$.

(3) When {\em v} is removed from {\em C}, {\em confChange}[v] = 0.

(4) When $\left \langle u,v \right \rangle$ is swapped, where {\em v} is removed from {\em C}, {\em confChange}[v] = 0.

In this intensification component, we propose a new prohibition mechanism named TabuCC, which is inspired by tabu strategy and SCC strategy. 
The aim of MVWCP is to find a clique with the maximum weight, therefore, intuitively, if a vertex {\em v} is added to the current clique, then its neighbors should also be encouraged to add to the clique \cite{WangCY2016}. Based on this idea, we propose a tabu-based prohibition mechanism. The TabuCC mechanism is worked as follows: 

i) For the {\em Swap} move, a vertex pair $\left \langle u,v \right \rangle \in V_{swap}(C)$, where {\em v} is removed from {\em C}, {\em v} is prohibited to be moved back to {\em C} for the next {\em tt\textunderscore swap} iterations.

{\em tt\textunderscore swap=random}$(|V_{swap}(C)|)+T_{1}$.

ii) For the {\em Drop} move, a vertex $u\in V_{drop}(C)$, where {\em u} is removed from {\em C}, {\em u} is prohibited to be moved back to {\em C} for the next {\em tt\textunderscore drop} iterations.

{\em tt\textunderscore drop}=$T_{1}$.

iii) For the {\em Add} move, a vertex $v\in V_{add}(C)$, where {\em v} is added into {\em C}, for each vertex $u\in N(v)$, lift the prohibition of {\em u}. 

The intensification component provides three prohibition mechanisms for selection, including the tabu mechanism proposed in MN/TS, SCC proposed in LSCC and TabuCC mentioned above. The tabu mechanism and TabuCC involve a parameter called {\em tabu\textunderscore tenure}, which is exposed to the configurator for selection. For selecting a vertex pair from $V_{swap}(C)$, {\em intensification\textunderscore process} procedure applies a fast and effective strategy named Best from Multiple Selection (BMS), which strikes a balance between quality and complexity and can bring diversity to the search process \cite{Cai2015}. The BMS strategy randomly selects {\em bms\textunderscore num} elements ({\em bms\textunderscore num} is an integer parameter) from source set {\em S}, and then returns the best element. The activation of BMS strategy is depended on a Boolean-valued parameter {\em perform\textunderscore BMS}. If BMS strategy is activated, the parameter {\em bms\textunderscore num} of BMS strategy will be activated.
There are three {\em selectDropVertexRule} in this component: i) Random selection; ii) Weight-based selection; iii) Perform random selection with a probability of {\em randomdrop\textunderscore prob}, otherwise perform weight-based selection. This component includes two {\em breakingTiesRules}: i) Breaking ties randomly; ii) Breaking ties in favor of the largest {\em age}.

\subsection {Configuration space and default configuration}
\label{subsec:cs_dc}
PbO-MWC is a parametric local search framework and can be configured to various high-performance local search algorithms. We have introduced the top level of our algorithm framework, all its components and the parameters in Subsections \ref{subsec:pbo-mwc-f}-\ref{subsec:intensification_c}. In Table \ref{tab:space}, we give an overview of the full configuration space of PbO-MWC, including all strategies and parameters, as well as the conditions under which strategies and parameters are activated. The default configuration settings of PbO-MWC are shown in Table \ref{tab:default}. 

\begin{table*} [pth]
\begin{center}
\caption{The optimized configurations of PbO-MWC for all benchmarks.}\label{tab:optimized}

\fontsize{\mytablefontsize}{\mytablebaselineskip}\selectfont\setlength{\tabcolsep}{\kestabcolsep}
\begin{spacing}{1}
\begin{tabular}{l l}

\hline
 
Benchmark/Instance Family & Optimized Configuration \\ 
\hline

BHOSLIB & {\em perform\textunderscore BMS}={\em False}, {\em breaking\textunderscore ties}=1, {\em init\textunderscore construction}=1, {\em drop\textunderscore vertex}=0, \\
 & {\em perform\textunderscore restart}={\em True}, {\em perform\textunderscore randomwalk}={\em True}, {\em restart\textunderscore prob}=5.016696977394702E-5, \\
 & {\em randomwalk\textunderscore prob}=0.09733547356349166, {\em tabu\textunderscore type}=1, {\em tabu\textunderscore tenure}=5  \\
& \\
 DIMACS & {\em perform\textunderscore BMS}={\em False}, {\em breaking\textunderscore ties}=1, {\em init\textunderscore construction}=1, {\em drop\textunderscore vertex}=1, \\
  (MANN family) & {\em perform\textunderscore restart}={\em False}, {\em perform\textunderscore randomwalk}={\em True}, {\em randomdrop\textunderscore prob}=0.1, \\
  & {\em randomwalk\textunderscore prob}=0.0021339029487367554, {\em tabu\textunderscore type}=0  \\
& \\
DIMACS & {\em perform\textunderscore BMS}={\em False}, {\em breaking\textunderscore ties}=1, {\em init\textunderscore construction}=0, {\em drop\textunderscore vertex}=0,\\
(except MANN family) & {\em perform\textunderscore restart}={\em True}, {\em perform\textunderscore randomwalk}={\em True}, {\em restart\textunderscore prob}=3.459685410644107E-5, \\
& {\em randomwalk\textunderscore prob}=0.00994485968433248, {\em tabu\textunderscore type}=1, {\em tabu\textunderscore tenure}=8  \\
& \\
KES & {\em perform\textunderscore BMS}={\em True}, {\em bms\textunderscore num}=6, {\em breaking\textunderscore ties}=1, {\em init\textunderscore construction}=0, {\em drop\textunderscore vertex}=2, \\
& {\em perform\textunderscore restart}={\em True}, {\em perform\textunderscore randomwalk}={\em False}, {\em restart\textunderscore prob}=2.7775287025690946E-5, \\
& {\em tabu\textunderscore type}=1, {\em tabu\textunderscore tenure}=30  \\
& \\
REF & {\em perform\textunderscore BMS}={\em True}, {\em bms\textunderscore num}=16, {\em breaking\textunderscore ties}=1, {\em init\textunderscore construction}=0, {\em drop\textunderscore vertex}=1, \\
& {\em perform\textunderscore restart}={\em True}, {\em perform\textunderscore randomwalk}={\em False}, {\em randomdrop\textunderscore prob}=0.4,\\
& {\em restart\textunderscore prob}=9.44211698679448E-6, {\em tabu\textunderscore type}=2, {\em tabu\textunderscore tenure}=8  \\

\hline

\end{tabular}
\end{spacing}

\end{center}
\end{table*}

\section{Experimental Evaluations}
\label{sec:experimental_evaluations}


To evaluate the efficiency of our proposed PbO-MWC framework and explore the potential of the configuration space,
we conduct extensive experiments to compare PbO-MWC against five state-of-the-art solvers on a broad range of MVWCP benchmarks. First, we describe the benchmarks and the competitors.
Second, we describe the configuration protocols used to automatically configure PbO-MWC and its competitors. Then we describe the experimental setup.
Finally, we present the experimental results and give some speculations about which strategies work well on which benchmark.

\subsection{The Benchmarks}
The set of 40 BHOSLIB instances arose from the SAT'04 Competition. The BHOSLIB instances were translated from hard random SAT instances. DIMACS benchmark set was established for the Second DIMACS Implementation Challenge. This set comprises 80 instances from a variety of real-world applications \cite{johnson1996cliques}. The BHOSLIB and DIMACS benchmarks have been widely used in the recent literature to test new MCP and MVWCP solvers \cite{richter2007stochastic,pullan2008approximating,pullan2011cooperating,wu2012multi,WangCY2016,fang2016exact,li2017minimization,fan2017restart}. The original graphs are unweighted, we adopt the method described in \cite{pullan2008approximating}: for each vertex {\em i}, $W_{i}$ is set to $(i\ mod\ 200) + 1$.

Besides the above two benchmarks, we evaluate the performance of all the solvers on two real-world application benchmarks. Kidney Exchange Scheme (KES) exists in several countries to increase the number of transplants from living donors to patients with end-stage renal disease. A donor-patient pair contains a patient and a person who is willing to donate to that patient but unable to do so due to non-compatible problem. Each feasible exchange gives a score that reflects its desirability. Typically, administrators perform matching operations at fixed intervals, with the goal of maximizing the sum of exchange scores. McCreesh et al. \cite{mccreesh2017maximum} proposed that this optimization problem may be solved by reduction to MVWCP, where each vertex is an exchange, whose weight is its score. Two exchanges are adjacent if and only if they have no participants in common. The clique stands for a maximally desirable set of donor-patient exchange. Research Excellence Framework (REF) is the system for assessing the quality of research in higher education institution. In each assessment unit, each staff would submit six publications, from which the organization chooses four. Cooperating authors within the same assessment unit cannot submit a shared publication. MVWCP helps the assessment units find a way to maximize the submission, where each vertex is a choice of four publications from six publications. The clique stands for the set of publications that an assessment unit can provide to the authority. These two benchmarks are generated by McCreesh et al. \cite{mccreesh2017maximum}. In this paper, we selected 42 instances the KES benchmark and 29 instances from the REF benchmark that are difficult to solve.

\begin{table*} [t]
\begin{center}
\caption{Experimental results on BHOSLIB benchmark. For all instances, each solver was performed with its optimized configuration trained on 5 instances in the upper part.}\label{tab:frb}
\fontsize{\mytablefontsize}{\mytablebaselineskip\baselineskip}\selectfont\setlength{\tabcolsep}{9pt}
\begin{tabular}{l  c c c c c c c}

\hline
 
\multirow{2}*{Graph}  & \multirow{2}*{{\em solBest}} & PbO-MWC & MN/TS & LSCC & LSCC+BMS & RRWL & TSM-MWC \\ 
	  &  & \multicolumn{1}{c}{{\em\#Suc}($t_{avg}$)} & \multicolumn{1}{c}{{\em\#Suc}($t_{avg}$)} & \multicolumn{1}{c}{{\em\#Suc}($t_{avg}$)} & \multicolumn{1}{c}{{\em\#Suc}($t_{avg}$)}& \multicolumn{1}{c}{{\em\#Suc}($t_{avg}$)} & \multicolumn{1}{c}{{\em\#Suc}({\em time})}\\
\hline\hline

frb45-21-1  &  4760  & \makecell[l]{\textbf{100}}\makecell[r]{(\textbf{14.459})} &  \makecell[l]{\textbf{100}}\makecell[r]{(68.809)} &  \makecell[l]{46}\makecell[r]{(1349.447)} &  \makecell[l]{42}\makecell[r]{(1414.568)} &  \makecell[l]{51}\makecell[r]{(1085.535)} &  \makecell[l]{0}\makecell[r]{(1029.450)} \\ 
frb45-21-2  &  4784  & \makecell[l]{\textbf{100}}\makecell[r]{(\textbf{1.759})} &  \makecell[l]{\textbf{100}}\makecell[r]{(14.099)} &  \makecell[l]{78}\makecell[r]{(1326.001)} &  \makecell[l]{70}\makecell[r]{(1352.767)} &  \makecell[l]{57}\makecell[r]{(1032.973)} &  \makecell[l]{0}\makecell[r]{(2733.750)} \\ 
frb45-21-3  &  4765  & \makecell[l]{\textbf{100}}\makecell[r]{(\textbf{2.565})} &  \makecell[l]{\textbf{100}}\makecell[r]{(22.378)} &  \makecell[l]{53}\makecell[r]{(1607.364)} &  \makecell[l]{51}\makecell[r]{(1638.062)} &  \makecell[l]{64}\makecell[r]{(1288.928)} &  \makecell[l]{0}\makecell[r]{(1563.770)} \\ 
frb45-21-4  &  4799  & \makecell[l]{\textbf{100}}\makecell[r]{(\textbf{1.245})} &  \makecell[l]{\textbf{100}}\makecell[r]{(49.663)} &  \makecell[l]{67}\makecell[r]{(1442.893)} &  \makecell[l]{59}\makecell[r]{(1510.165)} &  \makecell[l]{85}\makecell[r]{(1139.340)} &  \makecell[l]{0}\makecell[r]{(1329.360)} \\ 
frb45-21-5  &  4779  & \makecell[l]{\textbf{100}}\makecell[r]{(\textbf{2.294})} &  \makecell[l]{\textbf{100}}\makecell[r]{(5.323)} &  \makecell[l]{\textbf{100}}\makecell[r]{(301.269)} &  \makecell[l]{\textbf{100}}\makecell[r]{(338.229)} &  \makecell[l]{95}\makecell[r]{(393.415)} &  \makecell[l]{0}\makecell[r]{(1772.870)} \\ 
\hline

frb30-15-3  &  2995  & \makecell[l]{100}\makecell[r]{(\textbf{0.202})} &  \makecell[l]{100}\makecell[r]{(0.635)} &  \makecell[l]{100}\makecell[r]{(13.648)} &  \makecell[l]{100}\makecell[r]{(14.819)} &  \makecell[l]{100}\makecell[r]{(4.423)} &  \makecell[l]{100}\makecell[r]{(1562.590)} \\ 
frb35-17-1  &  3650  & \makecell[l]{\textbf{100}}\makecell[r]{(\textbf{0.808})} &  \makecell[l]{\textbf{100}}\makecell[r]{(4.672)} &  \makecell[l]{\textbf{100}}\makecell[r]{(91.153)} &  \makecell[l]{\textbf{100}}\makecell[r]{(107.303)} &  \makecell[l]{\textbf{100}}\makecell[r]{(38.994)} &  \makecell[l]{0}\makecell[r]{(3354.780)} \\ 
frb35-17-2  &  3738  & \makecell[l]{\textbf{100}}\makecell[r]{(\textbf{3.187})} &  \makecell[l]{\textbf{100}}\makecell[r]{(28.094)} &  \makecell[l]{\textbf{100}}\makecell[r]{(151.364)} &  \makecell[l]{\textbf{100}}\makecell[r]{(173.864)} &  \makecell[l]{\textbf{100}}\makecell[r]{(196.664)} &  \makecell[l]{0}\makecell[r]{(3276.140)} \\ 
frb35-17-3  &  3716  & \makecell[l]{\textbf{100}}\makecell[r]{(\textbf{0.378})} &  \makecell[l]{\textbf{100}}\makecell[r]{(4.365)} &  \makecell[l]{\textbf{100}}\makecell[r]{(42.298)} &  \makecell[l]{\textbf{100}}\makecell[r]{(49.195)} &  \makecell[l]{\textbf{100}}\makecell[r]{(29.877)} &  \makecell[l]{0}\makecell[r]{(2267.320)} \\ 
frb35-17-4  &  3683  & \makecell[l]{\textbf{100}}\makecell[r]{(\textbf{0.429})} &  \makecell[l]{\textbf{100}}\makecell[r]{(4.245)} &  \makecell[l]{\textbf{100}}\makecell[r]{(328.516)} &  \makecell[l]{\textbf{100}}\makecell[r]{(406.264)} &  \makecell[l]{\textbf{100}}\makecell[r]{(136.796)} &  \makecell[l]{0}\makecell[r]{(3440.670)} \\ 
frb35-17-5  &  3686  & \makecell[l]{\textbf{100}}\makecell[r]{(\textbf{0.548})} &  \makecell[l]{\textbf{100}}\makecell[r]{(1.058)} &  \makecell[l]{\textbf{100}}\makecell[r]{(20.187)} &  \makecell[l]{\textbf{100}}\makecell[r]{(22.571)} &  \makecell[l]{\textbf{100}}\makecell[r]{(19.197)} &  \makecell[l]{0}\makecell[r]{(2461.840)} \\ 
frb40-19-1  &  4063  & \makecell[l]{\textbf{100}}\makecell[r]{(\textbf{5.377})} &  \makecell[l]{\textbf{100}}\makecell[r]{(11.478)} &  \makecell[l]{\textbf{100}}\makecell[r]{(341.834)} &  \makecell[l]{\textbf{100}}\makecell[r]{(438.977)} &  \makecell[l]{\textbf{100}}\makecell[r]{(497.250)} &  \makecell[l]{0}\makecell[r]{(3344.290)} \\ 
frb40-19-2  &  4112  & \makecell[l]{\textbf{100}}\makecell[r]{(\textbf{2.137})} &  \makecell[l]{\textbf{100}}\makecell[r]{(23.773)} &  \makecell[l]{99}\makecell[r]{(656.502)} &  \makecell[l]{99}\makecell[r]{(747.700)} &  \makecell[l]{98}\makecell[r]{(879.786)} &  \makecell[l]{0}\makecell[r]{(3584.080)} \\ 
frb40-19-3  &  4115  & \makecell[l]{\textbf{100}}\makecell[r]{(\textbf{9.149})} &  \makecell[l]{\textbf{100}}\makecell[r]{(72.378)} &  \makecell[l]{97}\makecell[r]{(913.408)} &  \makecell[l]{94}\makecell[r]{(1017.261)} &  \makecell[l]{92}\makecell[r]{(898.788)} &  \makecell[l]{0}\makecell[r]{(2150.580)} \\ 
frb40-19-4  &  4136  & \makecell[l]{\textbf{100}}\makecell[r]{(\textbf{1.121})} &  \makecell[l]{\textbf{100}}\makecell[r]{(65.162)} &  \makecell[l]{97}\makecell[r]{(850.168)} &  \makecell[l]{96}\makecell[r]{(921.804)} &  \makecell[l]{95}\makecell[r]{(823.619)} &  \makecell[l]{0}\makecell[r]{(1722.160)} \\ 
frb40-19-5  &  4118  & \makecell[l]{\textbf{100}}\makecell[r]{(\textbf{2.928})} &  \makecell[l]{\textbf{100}}\makecell[r]{(16.579)} &  \makecell[l]{\textbf{100}}\makecell[r]{(434.633)} &  \makecell[l]{\textbf{100}}\makecell[r]{(514.314)} &  \makecell[l]{97}\makecell[r]{(317.954)} &  \makecell[l]{0}\makecell[r]{(1008.240)} \\ 
frb50-23-1  &  5494  & \makecell[l]{\textbf{100}}\makecell[r]{(\textbf{119.462})} &  \makecell[l]{77}\makecell[r]{(1532.625)} &  \makecell[l]{3}\makecell[r]{(1643.836)} &  \makecell[l]{1}\makecell[r]{(1741.232)} &  \makecell[l]{3}\makecell[r]{(1375.311)} &  \makecell[l]{0}\makecell[r]{(3040.930)} \\ 
frb50-23-2  &  5462  & \makecell[l]{\textbf{100}}\makecell[r]{(\textbf{92.994})} &  \makecell[l]{71}\makecell[r]{(1154.720)} &  \makecell[l]{6}\makecell[r]{(1577.357)} &  \makecell[l]{5}\makecell[r]{(1652.677)} &  \makecell[l]{10}\makecell[r]{(1372.736)} &  \makecell[l]{0}\makecell[r]{(319.980)} \\ 
frb50-23-3  &  5486  & \makecell[l]{\textbf{100}}\makecell[r]{(\textbf{17.533})} &  \makecell[l]{\textbf{100}}\makecell[r]{(47.560)} &  \makecell[l]{17}\makecell[r]{(1703.953)} &  \makecell[l]{11}\makecell[r]{(1619.835)} &  \makecell[l]{17}\makecell[r]{(1326.604)} &  \makecell[l]{0}\makecell[r]{(2949.470)} \\ 
frb50-23-4  &  5454  & \makecell[l]{\textbf{99}}\makecell[r]{(1035.311)} &  \makecell[l]{58}\makecell[r]{(919.783)} &  \makecell[l]{0}\makecell[r]{(1792.514)} &  \makecell[l]{0}\makecell[r]{(1729.724)} &  \makecell[l]{1}\makecell[r]{(1256.577)} &  \makecell[l]{0}\makecell[r]{(3560.910)} \\ 
frb50-23-5  &  5498  & \makecell[l]{\textbf{100}}\makecell[r]{(\textbf{5.399})} &  \makecell[l]{\textbf{100}}\makecell[r]{(26.746)} &  \makecell[l]{70}\makecell[r]{(1578.170)} &  \makecell[l]{62}\makecell[r]{(1653.499)} &  \makecell[l]{60}\makecell[r]{(1284.259)} &  \makecell[l]{0}\makecell[r]{(2748.320)} \\ 
frb53-24-1  &  5670  & \makecell[l]{\textbf{100}}\makecell[r]{(\textbf{33.530})} &  \makecell[l]{\textbf{100}}\makecell[r]{(398.036)} &  \makecell[l]{7}\makecell[r]{(1666.790)} &  \makecell[l]{3}\makecell[r]{(1788.058)} &  \makecell[l]{7}\makecell[r]{(1424.357)} &  \makecell[l]{0}\makecell[r]{(562.150)} \\ 
frb53-24-2  &  5707  & \makecell[l]{\textbf{100}}\makecell[r]{(\textbf{98.211})} &  \makecell[l]{54}\makecell[r]{(1681.134)} &  \makecell[l]{2}\makecell[r]{(1756.783)} &  \makecell[l]{3}\makecell[r]{(1668.641)} &  \makecell[l]{2}\makecell[r]{(1245.848)} &  \makecell[l]{0}\makecell[r]{(3453.760)} \\ 
frb53-24-3  &  5655  & \makecell[l]{\textbf{100}}\makecell[r]{(711.933)} &  \makecell[l]{29}\makecell[r]{(666.191)} &  \makecell[l]{1}\makecell[r]{(1836.278)} &  \makecell[l]{1}\makecell[r]{(1960.289)} &  \makecell[l]{0}\makecell[r]{(1511.852)} &  \makecell[l]{0}\makecell[r]{(3433.890)} \\ 
frb53-24-4  &  5714  & \makecell[l]{\textbf{100}}\makecell[r]{(\textbf{77.797})} &  \makecell[l]{30}\makecell[r]{(1051.233)} &  \makecell[l]{0}\makecell[r]{(1730.691)} &  \makecell[l]{0}\makecell[r]{(1801.601)} &  \makecell[l]{0}\makecell[r]{(1536.212)} &  \makecell[l]{0}\makecell[r]{(2215.230)} \\ 
frb53-24-5  &  5659  & \makecell[l]{\textbf{100}}\makecell[r]{(\textbf{286.630})} &  \makecell[l]{42}\makecell[r]{(1664.354)} &  \makecell[l]{0}\makecell[r]{(1641.510)} &  \makecell[l]{0}\makecell[r]{(1635.425)} &  \makecell[l]{1}\makecell[r]{(1314.714)} &  \makecell[l]{0}\makecell[r]{(1657.180)} \\ 
frb56-25-1  &  5916  & \makecell[l]{\textbf{100}}\makecell[r]{(\textbf{35.216})} &  \makecell[l]{97}\makecell[r]{(935.194)} &  \makecell[l]{1}\makecell[r]{(1996.991)} &  \makecell[l]{1}\makecell[r]{(1779.035)} &  \makecell[l]{2}\makecell[r]{(1459.226)} &  \makecell[l]{0}\makecell[r]{(3456.430)} \\ 
frb56-25-2  &  5886  & \makecell[l]{\textbf{100}}\makecell[r]{(\textbf{37.982})} &  \makecell[l]{35}\makecell[r]{(1727.162)} &  \makecell[l]{1}\makecell[r]{(1770.091)} &  \makecell[l]{0}\makecell[r]{(1798.876)} &  \makecell[l]{1}\makecell[r]{(1495.152)} &  \makecell[l]{0}\makecell[r]{(1467.520)} \\ 
frb56-25-3  &  5859  & \makecell[l]{\textbf{98}}\makecell[r]{(\textbf{706.840})} &  \makecell[l]{20}\makecell[r]{(1498.373)} &  \makecell[l]{0}\makecell[r]{(1868.919)} &  \makecell[l]{0}\makecell[r]{(1799.464)} &  \makecell[l]{0}\makecell[r]{(1461.475)} &  \makecell[l]{0}\makecell[r]{(2619.290)} \\ 
frb56-25-4  &  5892  & \makecell[l]{\textbf{99}}\makecell[r]{(\textbf{678.962})} &  \makecell[l]{30}\makecell[r]{(1719.638)} &  \makecell[l]{0}\makecell[r]{(1765.438)} &  \makecell[l]{0}\makecell[r]{(1862.983)} &  \makecell[l]{0}\makecell[r]{(1434.981)} &  \makecell[l]{0}\makecell[r]{(2525.100)} \\ 
frb56-25-5  &  5853  & \makecell[l]{\textbf{94}}\makecell[r]{(\textbf{1022.826})} &  \makecell[l]{12}\makecell[r]{(1594.286)} &  \makecell[l]{0}\makecell[r]{(1625.700)} &  \makecell[l]{0}\makecell[r]{(1730.110)} &  \makecell[l]{0}\makecell[r]{(1599.427)} &  \makecell[l]{0}\makecell[r]{(2425.130)} \\ 
frb59-26-1  &  6591  & \makecell[l]{\textbf{100}}\makecell[r]{(\textbf{33.607})} &  \makecell[l]{\textbf{100}}\makecell[r]{(464.091)} &  \makecell[l]{0}\makecell[r]{(1693.629)} &  \makecell[l]{0}\makecell[r]{(1737.817)} &  \makecell[l]{1}\makecell[r]{(1367.452)} &  \makecell[l]{0}\makecell[r]{(2930.000)} \\ 
frb59-26-2  &  6645  & \makecell[l]{\textbf{100}}\makecell[r]{(\textbf{111.328})} &  \makecell[l]{99}\makecell[r]{(632.977)} &  \makecell[l]{2}\makecell[r]{(1875.558)} &  \makecell[l]{0}\makecell[r]{(1843.966)} &  \makecell[l]{1}\makecell[r]{(1470.946)} &  \makecell[l]{0}\makecell[r]{(2124.560)} \\ 
frb59-26-3  &  6608  & \makecell[l]{\textbf{100}}\makecell[r]{(\textbf{74.141})} &  \makecell[l]{34}\makecell[r]{(1534.772)} &  \makecell[l]{0}\makecell[r]{(1991.458)} &  \makecell[l]{0}\makecell[r]{(1930.665)} &  \makecell[l]{0}\makecell[r]{(1513.631)} &  \makecell[l]{0}\makecell[r]{(1590.650)} \\ 
frb59-26-4  &  6592  & \makecell[l]{\textbf{100}}\makecell[r]{(92.463)} &  \makecell[l]{95}\makecell[r]{(988.683)} &  \makecell[l]{1}\makecell[r]{(1678.177)} &  \makecell[l]{1}\makecell[r]{(1727.378)} &  \makecell[l]{1}\makecell[r]{(1502.068)} &  \makecell[l]{0}\makecell[r]{(23.280)} \\ 
frb59-26-5  &  6584  & \makecell[l]{\textbf{100}}\makecell[r]{(\textbf{279.545})} &  \makecell[l]{48}\makecell[r]{(1429.216)} &  \makecell[l]{0}\makecell[r]{(1652.560)} &  \makecell[l]{0}\makecell[r]{(1797.136)} &  \makecell[l]{0}\makecell[r]{(1536.050)} &  \makecell[l]{0}\makecell[r]{(2806.550)} \\ 
 
 \hline
\end{tabular}

\end{center}
\end{table*}

\subsection{Competitors}
In this paper, we compare PbO-MWC (the implementation is available online.\footnote{https://github.com/PbO-MWC/PbO-MWC}) against five state-of-the-art solvers, including four SLS solvers and one complete solver:

\textbf{SLS solvers:}

MN/TS\cite{wu2012multi} is a high performance SLS solver based on multi-neighborhood search and tabu mechanism. In our experiments, we used the version of MN/TS made available by its authors.\footnote{http://www.info.univ-angers.fr/\%7ehao/clique.html}

LSCC\cite{WangCY2016} is an efficient SLS solver which perform well on BHOSLIB and DIMACS. LSCC+BMS\cite{WangCY2016} is suitable for massive graph instances. We used the version of these two solvers that are available online.\footnote{http://ai.nenu.edu.cn/wangyy/Yiyuandata/Download/mwcp.tar.gz}

RRWL\cite{fan2017restart} is an efficient SLS solver without parameters. We used the version of RRWL made available by its authors.\footnote{https://github.com/Fan-Yi/Restart-and-Random-Walk-in-Local-Search-for-MVWC-in-Large-Sparse-Graphs}

\textbf{Complete solver:}

TSM-MWC\cite{jiang2018two} is a state-of-the-art complete solver for MWC in both small/medium and massive real-world graphs. The source code is available online.\footnote{https://home.mis.u-picardie.fr/\%7ecli/EnglishPage.html}

\subsection{Configuration Protocol}
In this work, we made use of SMAC(version:2.10.03) to automatically configure our PbO-MWC framework. In this subsection, we describe the protocol used for PbO-MWC and its competitors. We extracted a training set for each benchmark. For BHOSLIB, {\em frb45} family is chosen to be the training set. For DIMACS and REF, we randomly chosen an instance from each family. For KES, we randomly chosen 10 instances from the benchmark.
For some training sets, SMAC could not find a configuration for some solvers, with the configuration, the solver could reach the known optimal solution at least once within a cutoff time for each run. In order to configure all the solvers in a uniform protocol, we define a new solution quality named {\em NewSQ}: {\em NewSQ}=0-({\em solution quality}) + ({\em solution time})/1000. We used SMAC to minimize {\em NewSQ}. Throughout the configuration process, we allowed a 2-day time budget and a cutoff time of 300 CPU seconds for each solver run. For each training set, we performed 25 SMAC runs and obtained 25 optimized configurations. The configuration with the minimum {\em NewSQ} value on the training set was chosen as the final optimized configuration. The optimized configurations of PbO-MWC for each benchmark are shown in Table \ref{tab:optimized}.

\begin{table*} [t]
\begin{center}
\caption{Experimental results on DIMACS benchmark. For the {\em MANN} instance family, each solver was performed with its optimized configuration trained on MANN\textunderscore a45 instance. For other instances, each solver was performed with its optimized configuration trained on 9 instances of the instances in the upper part that do not contain MANN\textunderscore a45.}\label{tab:dimacs}
\fontsize{\mytablefontsize}{\mytablebaselineskip\baselineskip}\selectfont\setlength{\tabcolsep}{\kestabcolsep}

\resizebox{\textwidth}{!}{
\begin{tabular}{l  c c c c c c }

\hline
 
\multirow{3}*{Graph}  & PbO-MWC & MN/TS & LSCC & LSCC+BMS & RRWL & TSM-MWC \\ 
	  & \multicolumn{1}{l}{$w_{max}$($w_{avg}$)} & \multicolumn{1}{l}{$w_{max}$($w_{avg}$)} & \multicolumn{1}{l}{$w_{max}$($w_{avg}$)} & \multicolumn{1}{l}{$w_{max}$($w_{avg}$)}& \multicolumn{1}{l}{$w_{max}$($w_{avg}$)} & \multicolumn{1}{l}{$w_{sol}$}\\
	  & \multicolumn{1}{r}{$t_{avg}$} & \multicolumn{1}{r}{$t_{avg}$} & \multicolumn{1}{r}{$t_{avg}$} & \multicolumn{1}{r}{$t_{avg}$}& \multicolumn{1}{r}{$t_{avg}$}& 
	  \multicolumn{1}{r}{{\em time}}\\
	  
\hline\hline
\multirow{2}*{MANN\textunderscore a45} & \makecell[l]{34263(34262.59)} & \makecell[l]{34226(34199.31)} & \makecell[l]{34256(34254.02)} & \makecell[l]{34258(34253.84)} & \makecell[l]{34263(34254.72)} & \makecell[l]{\textbf{34265}} \\ 
 & \makecell[r]{1490.467} & \makecell[r]{1815.412} & \makecell[r]{425.650} & \makecell[r]{1291.260} & \makecell[r]{357.249} &  \makecell[r]{404.800}  \\ 
\multirow{2}*{brock800\textunderscore 4} & \makecell[l]{2971(\textbf{2971.00})} & \makecell[l]{2971(2970.98)} & \makecell[l]{2971(2970.80)} & \makecell[l]{2971(2970.78)} & \makecell[l]{2971(\textbf{2971.00})} & \makecell[l]{2971} \\ 
 & \makecell[r]{\textbf{42.734}} & \makecell[r]{774.713} & \makecell[r]{1176.934} & \makecell[r]{1128.025} & \makecell[r]{126.596} &  \makecell[r]{2540.720}  \\ 
\multirow{2}*{C2000.9} & \makecell[l]{\textbf{10999}(\textbf{10999.00})} & \makecell[l]{\textbf{10999}(\textbf{10999.00})} & \makecell[l]{\textbf{10999}(10951.90)} & \makecell[l]{\textbf{10999}(10951.25)} & \makecell[l]{\textbf{10999}(10951.41)} & \makecell[l]{8338} \\ 
 & \makecell[r]{\textbf{101.025}} & \makecell[r]{191.816} & \makecell[r]{1919.433} & \makecell[r]{1902.930} & \makecell[r]{1437.638} &  \makecell[r]{2311.820}  \\ 
\multirow{2}*{c-fat500-10} & \makecell[l]{11586(11586.00)} & \makecell[l]{11586(11586.00)} & \makecell[l]{11586(11586.00)} & \makecell[l]{11586(11586.00)} & \makecell[l]{11586(11586.00)} & \makecell[l]{11586} \\ 
 & \makecell[r]{0.248} & \makecell[r]{0.059} & \makecell[r]{\textbf{\textless 0.001}} & \makecell[r]{\textbf{\textless 0.001}} & \makecell[r]{0.379} &  \makecell[r]{0.190}  \\ 
\multirow{2}*{DSJC1000.5} & \makecell[l]{2186(2186.00)} & \makecell[l]{2186(2186.00)} & \makecell[l]{2186(2186.00)} & \makecell[l]{2186(2186.00)} & \makecell[l]{2186(2186.00)} & \makecell[l]{2186} \\ 
 & \makecell[r]{0.083} & \makecell[r]{\textbf{0.047}} & \makecell[r]{5.955} & \makecell[r]{5.989} & \makecell[r]{1.158} &  \makecell[r]{54.910}  \\ 
\multirow{2}*{gen400\textunderscore p0.9\textunderscore 75} & \makecell[l]{8006(8006.00)} & \makecell[l]{8006(8006.00)} & \makecell[l]{8006(8006.00)} & \makecell[l]{8006(8006.00)} & \makecell[l]{8006(8006.00)} & \makecell[l]{8006} \\ 
 & \makecell[r]{\textbf{0.001}} & \makecell[r]{0.007} & \makecell[r]{0.638} & \makecell[r]{0.693} & \makecell[r]{0.538} &  \makecell[r]{77.200}  \\ 
\multirow{2}*{hamming10-2} & \makecell[l]{50512(50512.00)} & \makecell[l]{50512(50512.00)} & \makecell[l]{50512(50512.00)} & \makecell[l]{50512(50512.00)} & \makecell[l]{50512(50512.00)} & \makecell[l]{50512} \\ 
 & \makecell[r]{\textbf{0.145}} & \makecell[r]{0.652} & \makecell[r]{0.588} & \makecell[r]{0.516} & \makecell[r]{0.966} &  \makecell[r]{43.290}  \\ 
\multirow{2}*{johnson32-2-4} & \makecell[l]{\textbf{2033}(\textbf{2033.00})} & \makecell[l]{\textbf{2033}(\textbf{2033.00})} & \makecell[l]{\textbf{2033}(\textbf{2033.00})} & \makecell[l]{\textbf{2033}(\textbf{2033.00})} & \makecell[l]{\textbf{2033}(\textbf{2033.00})} & \makecell[l]{1891} \\ 
 & \makecell[r]{\textbf{0.003}} & \makecell[r]{0.811} & \makecell[r]{0.151} & \makecell[r]{0.154} & \makecell[r]{0.410} &  \makecell[r]{10.590}  \\ 
\multirow{2}*{p\textunderscore hat1500-3} & \makecell[l]{10321(10321.00)} & \makecell[l]{10321(10321.00)} & \makecell[l]{10321(10321.00)} & \makecell[l]{10321(10321.00)} & \makecell[l]{10321(10321.00)} & \makecell[l]{10321} \\ 
 & \makecell[r]{\textbf{2.855}} & \makecell[r]{29.639} & \makecell[r]{113.621} & \makecell[r]{117.621} & \makecell[r]{30.924} &  \makecell[r]{3336.320}  \\ 
\multirow{2}*{san400\textunderscore 0.9\textunderscore 1} & \makecell[l]{9776(9776.00)} & \makecell[l]{9776(9776.00)} & \makecell[l]{9776(9776.00)} & \makecell[l]{9776(9776.00)} & \makecell[l]{9776(9776.00)} & \makecell[l]{9776} \\ 
 & \makecell[r]{3.402} & \makecell[r]{\textbf{1.646}} & \makecell[r]{2.848} & \makecell[r]{3.218} & \makecell[r]{3.511} &  \makecell[r]{75.290}  \\ 
 
\hline
\multirow{2}*{MANN\textunderscore a27} & \makecell[l]{\textbf{12283}(\textbf{12283.00})} & \makecell[l]{12282(12276.98)} & \makecell[l]{\textbf{12283}(\textbf{12283.00})} & \makecell[l]{\textbf{12283}(\textbf{12283.00})} & \makecell[l]{\textbf{12283}(\textbf{12283.00})} & \makecell[l]{\textbf{12283}} \\ 
 & \makecell[r]{\textbf{3.974}} & \makecell[r]{1377.077} & \makecell[r]{129.249} & \makecell[r]{251.788} & \makecell[r]{270.134} &  \makecell[r]{4.400}  \\ 
\multirow{2}*{MANN\textunderscore a81} & \makecell[l]{\textbf{111355}(\textbf{111342.37})} & \makecell[l]{110171(110090.74)} & \makecell[l]{111302(111250.54)} & \makecell[l]{111269(111207.88)} & \makecell[l]{111324(111303.34)} & \makecell[l]{109970} \\ 
 & \makecell[r]{1639.896} & \makecell[r]{1818.422} & \makecell[r]{1639.686} & \makecell[r]{1861.101} & \makecell[r]{1784.362} &  \makecell[r]{3202.890}  \\ 
\multirow{2}*{brock400\textunderscore 4} & \makecell[l]{3626(3626.00)} & \makecell[l]{3626(3626.00)} & \makecell[l]{3626(3626.00)} & \makecell[l]{3626(3626.00)} & \makecell[l]{3626(3626.00)} & \makecell[l]{3626} \\ 
 & \makecell[r]{\textbf{0.632}} & \makecell[r]{0.988} & \makecell[r]{12.447} & \makecell[r]{13.083} & \makecell[r]{1.634} &  \makecell[r]{136.900}  \\ 
\multirow{2}*{C1000.9} & \makecell[l]{\textbf{9254}(\textbf{9254.00})} & \makecell[l]{\textbf{9254}(\textbf{9254.00})} & \makecell[l]{\textbf{9254}(\textbf{9254.00})} & \makecell[l]{\textbf{9254}(\textbf{9254.00})} & \makecell[l]{\textbf{9254}(\textbf{9254.00})} & \makecell[l]{7477} \\ 
 & \makecell[r]{1.214} & \makecell[r]{\textbf{1.201}} & \makecell[r]{177.922} & \makecell[r]{186.512} & \makecell[r]{63.886} &  \makecell[r]{2806.730}  \\ 
\multirow{2}*{C4000.5} & \makecell[l]{\textbf{2792}(\textbf{2792.00})} & \makecell[l]{\textbf{2792}(\textbf{2792.00})} & \makecell[l]{\textbf{2792}(\textbf{2792.00})} & \makecell[l]{\textbf{2792}(\textbf{2792.00})} & \makecell[l]{\textbf{2792}(\textbf{2792.00})} & \makecell[l]{2502} \\ 
 & \makecell[r]{\textbf{13.808}} & \makecell[r]{14.050} & \makecell[r]{77.724} & \makecell[r]{79.793} & \makecell[r]{129.973} &  \makecell[r]{3497.290}  \\ 
\multirow{2}*{hamming10-4} & \makecell[l]{\textbf{5129}(\textbf{5129.00})} & \makecell[l]{\textbf{5129}(\textbf{5129.00})} & \makecell[l]{\textbf{5129}(\textbf{5129.00})} & \makecell[l]{\textbf{5129}(\textbf{5129.00})} & \makecell[l]{\textbf{5129}(\textbf{5129.00})} & \makecell[l]{4828} \\ 
 & \makecell[r]{\textbf{1.148}} & \makecell[r]{2.846} & \makecell[r]{19.140} & \makecell[r]{21.761} & \makecell[r]{23.739} &  \makecell[r]{1244.040}  \\ 
\multirow{2}*{keller5} & \makecell[l]{\textbf{3317}(\textbf{3317.00})} & \makecell[l]{\textbf{3317}(\textbf{3317.00})} & \makecell[l]{\textbf{3317}(\textbf{3317.00})} & \makecell[l]{\textbf{3317}(\textbf{3317.00})} & \makecell[l]{\textbf{3317}(\textbf{3317.00})} & \makecell[l]{3097} \\ 
 & \makecell[r]{0.332} & \makecell[r]{\textbf{0.245}} & \makecell[r]{15.977} & \makecell[r]{18.548} & \makecell[r]{6.088} &  \makecell[r]{3472.040}  \\ 
\multirow{2}*{keller6} & \makecell[l]{\textbf{8062}(\textbf{8062.00})} & \makecell[l]{\textbf{8062}(\textbf{8062.00})} & \makecell[l]{\textbf{8062}(7858.60)} & \makecell[l]{\textbf{8062}(7862.85)} & \makecell[l]{\textbf{8062}(7892.65)} & \makecell[l]{4793} \\ 
 & \makecell[r]{\textbf{83.535}} & \makecell[r]{509.848} & \makecell[r]{1729.344} & \makecell[r]{1895.647} & \makecell[r]{1633.382} &  \makecell[r]{3564.280}  \\ 
\multirow{2}*{san1000} & \makecell[l]{1716(1716.00)} & \makecell[l]{1716(1716.00)} & \makecell[l]{1716(1716.00)} & \makecell[l]{1716(1716.00)} & \makecell[l]{1716(1716.00)} & \makecell[l]{1716} \\ 
 & \makecell[r]{11.055} & \makecell[r]{7.871} & \makecell[r]{656.871} & \makecell[r]{90.903} & \makecell[r]{14.471} &  \makecell[r]{\textbf{5.580}}  \\ 
\multirow{2}*{san400\textunderscore 0.7\textunderscore 1} & \makecell[l]{3941(3941.00)} & \makecell[l]{3941(3941.00)} & \makecell[l]{3941(3941.00)} & \makecell[l]{3941(3941.00)} & \makecell[l]{3941(3941.00)} & \makecell[l]{3941} \\ 
 & \makecell[r]{158.120} & \makecell[r]{42.184} & \makecell[r]{62.246} & \makecell[r]{88.384} & \makecell[r]{7.087} &  \makecell[r]{\textbf{1.490}}  \\ 
\multirow{2}*{san400\textunderscore 0.7\textunderscore 2} & \makecell[l]{3110(3110.00)} & \makecell[l]{3110(3110.00)} & \makecell[l]{3110(3110.00)} & \makecell[l]{3110(3110.00)} & \makecell[l]{3110(3110.00)} & \makecell[l]{3110} \\ 
 & \makecell[r]{245.999} & \makecell[r]{75.019} & \makecell[r]{197.050} & \makecell[r]{223.260} & \makecell[r]{15.661} &  \makecell[r]{\textbf{3.890}}  \\  
 
 \hline
\end{tabular}
}

\end{center}
\end{table*}

\subsection{Experimental Setup}
All the experiments were carried out on a workstation under the operating system CentOS (version: 7.6.1810), with Intel(R) Xeon(R) CPU E5-2620 2.10GHz CPU, 20MB L3 cache and 128GB RAM. Except RRWL and TSM-MWC without parameters, other solvers were configured using SMAC with the same configuration protocol. Each local search solver was executed 100 runs on each instance with seeds from 1 to 100. TSM-MWC was performed 1 run on each instance. The cutoff time for each solver run was set to 3600 seconds. The best solution-quality known so far is indicated by {\em solBest}. We report the number of successful runs (reaches {\em solBest} within cutoff time), denoted by {\em\#Suc}, and the averaged running time of finding the final solution on each instance, denoted by $t_{avg}$. We denote the number of successful runs divided by the number of total runs as {\em success rate}. Since the main difference between solvers on DIMACS is the solution quality, for each solver on each instance of DIMACS, on the 100 runs, we report the maximum weight ($w_{max}$) and averaged weight ($w_{avg}$) of the cliques found by each solver. For TSM-MWC, we report the weight of final clique found, denoted by $w_{sol}$, and the time of the final clique found, denoted by {\em time}. We do not report the instances that all the local search solvers reach 100\% {\em success rate} with $t_{avg}$\textless 10 seconds. In this experiments, unspecified time units are \textbf{CPU seconds}.

We also report the averaged PAR10 (penalized running time, if the solver can not get the {\em solBest} in a given cutoff time, it counts the running time as 10 times the given cutoff time.) of each solver on each benchmark, denoted by {\em avgPAR10}.

\subsection{Experimental Results}



\subsubsection {Results on BHOSLIB Benchmark.}
Table \ref{tab:frb} presents the comparative results of PbO-MWC and its competitors on BHOSLIB benchmark. From Table \ref{tab:frb}, we can clearly see that our PbO-MWC algorithm stands out the best solver on the training set and test set.
\textbf{On training set}, PbO-MWC achieves 100\% {\em success rate} on all 5 instances. MN/TS, the second best solver also achieves a 100\% {\em success rate} on all training instances, but in terms of running time, the averaged time of PbO-MWC is 4.464 and the averaged time of MN/TS is 32.054. 
The {\em success rate} of LSCC, LSCC+BMS and RRWL are much lower than 100\%. TSM-MWC can not reach the best solution known on any training instance. \textbf{On all 31 test instances}, PbO-MWC achieves a 100\% {\em success rate} for 27 of them, while the figure is 15, 8, 8, 7 and 1 for MN/TS, LSCC, LSCC+BMS, RRWL and TSM-MWC, respectively. On the four instances where PbO-MWC can't reach the 100\% {\em success rate}, the {\em success rate} of PbO-MWC is 99\%, 98\%, 99\% and 94\% respectively, which is much higher than that of its competitors. The total {\em\#Suc} of PbO-MWC is 3090, while the figure is 2331, 1264, 1177, 1189 and 100 for its competitors respectively. In addition, PbO-MWC finds best solutions with shortest averaged time on 28 of 31 instances. The averaged time of PbO-MWC is 179.935, the figure is 706.426, 1248.068, 1279.758, 1042.975 and 2389.775 for MN/TS, LSCC, LSCC+BMS, RRWL and TSM-MWC respectively.

\begin{table*} [t]
\begin{center}
\caption{Experimental results on KES benchmark. For all instances, each solver was performed with its optimized configuration trained on 10 instances in the upper part.}\label{tab:kes}
\fontsize{\mytablefontsize}{\mytablebaselineskip\baselineskip}\selectfont\setlength{\tabcolsep}{\kestabcolsep}
\begin{tabular}{l  c c c c c c c}

\hline
 
\multirow{2}*{Graph}  & \multirow{2}*{{\em solBest}} & PbO-MWC & MN/TS & LSCC & LSCC+BMS & RRWL & TSM-MWC \\ 
	  &  & \multicolumn{1}{c}{{\em\#Suc}($t_{avg}$)} & \multicolumn{1}{c}{{\em\#Suc}($t_{avg}$)} & \multicolumn{1}{c}{{\em\#Suc}($t_{avg}$)} & \multicolumn{1}{c}{{\em\#Suc}($t_{avg}$)}& \multicolumn{1}{c}{{\em\#Suc}($t_{avg}$)} & 
	  \multicolumn{1}{c}{{\em\#Suc}({\em time})}\\
\hline\hline

83  &  1237688860685  & \makecell[l]{100}\makecell[r]{(\textbf{0.895})} &  \makecell[l]{100}\makecell[r]{(76.022)} &  \makecell[l]{100}\makecell[r]{(73.769)} &  \makecell[l]{100}\makecell[r]{(93.026)} &  \makecell[l]{100}\makecell[r]{(142.617)} &  \makecell[l]{100}\makecell[r]{(162.390)} \\ 
84  &  1100166012937  & \makecell[l]{100}\makecell[r]{(\textbf{0.593})} &  \makecell[l]{100}\makecell[r]{(170.961)} &  \makecell[l]{100}\makecell[r]{(5.882)} &  \makecell[l]{100}\makecell[r]{(12.575)} &  \makecell[l]{100}\makecell[r]{(10.478)} &  \makecell[l]{100}\makecell[r]{(189.200)} \\ 
91  &  1306441900046  & \makecell[l]{\textbf{100}}\makecell[r]{(\textbf{4.447})} &  \makecell[l]{90}\makecell[r]{(1274.412)} &  \makecell[l]{\textbf{100}}\makecell[r]{(267.249)} &  \makecell[l]{86}\makecell[r]{(1148.002)} &  \makecell[l]{\textbf{100}}\makecell[r]{(348.165)} &  \makecell[l]{0}\makecell[r]{(112.590)} \\ 
96  &  1375094325251  & \makecell[l]{\textbf{100}}\makecell[r]{(\textbf{1.572})} &  \makecell[l]{13}\makecell[r]{(271.969)} &  \makecell[l]{\textbf{100}}\makecell[r]{(272.134)} &  \makecell[l]{85}\makecell[r]{(959.731)} &  \makecell[l]{\textbf{100}}\makecell[r]{(440.349)} &  \makecell[l]{0}\makecell[r]{(6.290)} \\ 
97  &  1375144632330  & \makecell[l]{\textbf{100}}\makecell[r]{(\textbf{2.045})} &  \makecell[l]{3}\makecell[r]{(507.525)} &  \makecell[l]{\textbf{100}}\makecell[r]{(414.291)} &  \makecell[l]{82}\makecell[r]{(1196.935)} &  \makecell[l]{73}\makecell[r]{(1052.616)} &  \makecell[l]{0}\makecell[r]{(744.600)} \\ 
105  &  1787797020693  & \makecell[l]{\textbf{100}}\makecell[r]{(\textbf{7.879})} &  \makecell[l]{92}\makecell[r]{(1073.198)} &  \makecell[l]{74}\makecell[r]{(1363.477)} &  \makecell[l]{70}\makecell[r]{(1293.008)} &  \makecell[l]{43}\makecell[r]{(1177.404)} &  \makecell[l]{0}\makecell[r]{(119.590)} \\ 
107  &  1650240659468  & \makecell[l]{\textbf{100}}\makecell[r]{(\textbf{4.792})} &  \makecell[l]{95}\makecell[r]{(1011.776)} &  \makecell[l]{78}\makecell[r]{(1196.789)} &  \makecell[l]{52}\makecell[r]{(1548.089)} &  \makecell[l]{41}\makecell[r]{(1312.153)} &  \makecell[l]{0}\makecell[r]{(148.390)} \\ 
114  &  2406557630478  & \makecell[l]{\textbf{100}}\makecell[r]{(\textbf{17.364})} &  \makecell[l]{0}\makecell[r]{(1535.306)} &  \makecell[l]{2}\makecell[r]{(1590.658)} &  \makecell[l]{2}\makecell[r]{(1693.137)} &  \makecell[l]{2}\makecell[r]{(1776.499)} &  \makecell[l]{0}\makecell[r]{(1671.880)} \\ 
116  &  2131511910416  & \makecell[l]{\textbf{100}}\makecell[r]{(\textbf{5.393})} &  \makecell[l]{0}\makecell[r]{(1383.749)} &  \makecell[l]{82}\makecell[r]{(1110.178)} &  \makecell[l]{68}\makecell[r]{(1349.423)} &  \makecell[l]{38}\makecell[r]{(1491.430)} &  \makecell[l]{0}\makecell[r]{(414.780)} \\ 
119  &  2269068296209  & \makecell[l]{\textbf{100}}\makecell[r]{(\textbf{4.812})} &  \makecell[l]{3}\makecell[r]{(1220.001)} &  \makecell[l]{\textbf{100}}\makecell[r]{(679.383)} &  \makecell[l]{99}\makecell[r]{(743.593)} &  \makecell[l]{75}\makecell[r]{(1281.739)} &  \makecell[l]{0}\makecell[r]{(488.960)} \\ 

\hline

71  &  1306458693642  & \makecell[l]{100}\makecell[r]{(\textbf{0.307})} &  \makecell[l]{100}\makecell[r]{(0.506)} &  \makecell[l]{100}\makecell[r]{(3.692)} &  \makecell[l]{100}\makecell[r]{(2.532)} &  \makecell[l]{100}\makecell[r]{(10.101)} &  \makecell[l]{100}\makecell[r]{(194.990)} \\ 
81  &  1650240634895  & \makecell[l]{\textbf{100}}\makecell[r]{(\textbf{10.112})} &  \makecell[l]{42}\makecell[r]{(952.869)} &  \makecell[l]{92}\makecell[r]{(928.713)} &  \makecell[l]{98}\makecell[r]{(1004.306)} &  \makecell[l]{51}\makecell[r]{(1255.399)} &  \makecell[l]{0}\makecell[r]{(994.070)} \\ 
82  &  1443914440714  & \makecell[l]{\textbf{100}}\makecell[r]{(\textbf{2.676})} &  \makecell[l]{0}\makecell[r]{(1454.161)} &  \makecell[l]{\textbf{100}}\makecell[r]{(52.113)} &  \makecell[l]{\textbf{100}}\makecell[r]{(52.100)} &  \makecell[l]{\textbf{100}}\makecell[r]{(96.197)} &  \makecell[l]{\textbf{100}}\makecell[r]{(601.480)} \\ 
85  &  1100216320013  & \makecell[l]{100}\makecell[r]{(\textbf{1.168})} &  \makecell[l]{100}\makecell[r]{(3.232)} &  \makecell[l]{100}\makecell[r]{(33.098)} &  \makecell[l]{100}\makecell[r]{(63.250)} &  \makecell[l]{100}\makecell[r]{(71.106)} &  \makecell[l]{100}\makecell[r]{(14.990)} \\ 
87  &  1375194939405  & \makecell[l]{100}\makecell[r]{(\textbf{1.127})} &  \makecell[l]{100}\makecell[r]{(3.583)} &  \makecell[l]{100}\makecell[r]{(50.560)} &  \makecell[l]{100}\makecell[r]{(34.520)} &  \makecell[l]{100}\makecell[r]{(171.926)} &  \makecell[l]{100}\makecell[r]{(8.060)} \\ 
89  &  893940457482  & \makecell[l]{100}\makecell[r]{(\textbf{0.329})} &  \makecell[l]{100}\makecell[r]{(7.763)} &  \makecell[l]{100}\makecell[r]{(59.544)} &  \makecell[l]{100}\makecell[r]{(52.238)} &  \makecell[l]{100}\makecell[r]{(75.254)} &  \makecell[l]{100}\makecell[r]{(0.790)} \\ 
92  &  1581403750408  & \makecell[l]{\textbf{100}}\makecell[r]{(\textbf{1.738})} &  \makecell[l]{0}\makecell[r]{(1159.796)} &  \makecell[l]{\textbf{100}}\makecell[r]{(544.064)} &  \makecell[l]{93}\makecell[r]{(1182.126)} &  \makecell[l]{85}\makecell[r]{(1183.945)} &  \makecell[l]{\textbf{100}}\makecell[r]{(5.690)} \\ 
94  &  1031547150353  & \makecell[l]{100}\makecell[r]{(\textbf{1.118})} &  \makecell[l]{100}\makecell[r]{(2.025)} &  \makecell[l]{100}\makecell[r]{(106.281)} &  \makecell[l]{100}\makecell[r]{(49.779)} &  \makecell[l]{100}\makecell[r]{(137.152)} &  \makecell[l]{100}\makecell[r]{(77.780)} \\ 
95  &  1375245246480  & \makecell[l]{\textbf{100}}\makecell[r]{(\textbf{2.556})} &  \makecell[l]{\textbf{100}}\makecell[r]{(48.995)} &  \makecell[l]{\textbf{100}}\makecell[r]{(192.375)} &  \makecell[l]{\textbf{100}}\makecell[r]{(432.595)} &  \makecell[l]{\textbf{100}}\makecell[r]{(418.615)} &  \makecell[l]{0}\makecell[r]{(14.880)} \\ 
98  &  1443947978765  & \makecell[l]{100}\makecell[r]{(\textbf{1.156})} &  \makecell[l]{100}\makecell[r]{(8.000)} &  \makecell[l]{100}\makecell[r]{(13.263)} &  \makecell[l]{100}\makecell[r]{(11.132)} &  \makecell[l]{100}\makecell[r]{(21.734)} &  \makecell[l]{100}\makecell[r]{(501.780)} \\ 
99  &  1237722398735  & \makecell[l]{\textbf{100}}\makecell[r]{(\textbf{1.985})} &  \makecell[l]{\textbf{100}}\makecell[r]{(11.295)} &  \makecell[l]{\textbf{100}}\makecell[r]{(18.543)} &  \makecell[l]{\textbf{100}}\makecell[r]{(19.415)} &  \makecell[l]{\textbf{100}}\makecell[r]{(38.765)} &  \makecell[l]{0}\makecell[r]{(974.380)} \\ 
100  &  1512701018124  & \makecell[l]{\textbf{100}}\makecell[r]{(\textbf{1.586})} &  \makecell[l]{21}\makecell[r]{(524.223)} &  \makecell[l]{\textbf{100}}\makecell[r]{(185.291)} &  \makecell[l]{\textbf{100}}\makecell[r]{(230.710)} &  \makecell[l]{\textbf{100}}\makecell[r]{(515.888)} &  \makecell[l]{0}\makecell[r]{(13.090)} \\ 
101  &  1650207096842  & \makecell[l]{\textbf{100}}\makecell[r]{(\textbf{5.288})} &  \makecell[l]{0}\makecell[r]{(1246.877)} &  \makecell[l]{83}\makecell[r]{(1174.490)} &  \makecell[l]{23}\makecell[r]{(1066.035)} &  \makecell[l]{68}\makecell[r]{(1473.039)} &  \makecell[l]{\textbf{100}}\makecell[r]{(20.390)} \\ 
102  &  1718960136202  & \makecell[l]{\textbf{100}}\makecell[r]{(\textbf{2.812})} &  \makecell[l]{30}\makecell[r]{(623.679)} &  \makecell[l]{\textbf{100}}\makecell[r]{(92.241)} &  \makecell[l]{95}\makecell[r]{(824.607)} &  \makecell[l]{\textbf{100}}\makecell[r]{(236.558)} &  \makecell[l]{0}\makecell[r]{(132.650)} \\ 
103  &  1512701018125  & \makecell[l]{\textbf{100}}\makecell[r]{(\textbf{3.859})} &  \makecell[l]{38}\makecell[r]{(668.578)} &  \makecell[l]{\textbf{100}}\makecell[r]{(62.515)} &  \makecell[l]{\textbf{100}}\makecell[r]{(109.763)} &  \makecell[l]{\textbf{100}}\makecell[r]{(82.190)} &  \makecell[l]{0}\makecell[r]{(34.780)} \\ 
104  &  1512818425875  & \makecell[l]{\textbf{100}}\makecell[r]{(\textbf{4.727})} &  \makecell[l]{40}\makecell[r]{(1647.883)} &  \makecell[l]{78}\makecell[r]{(1141.392)} &  \makecell[l]{74}\makecell[r]{(1244.486)} &  \makecell[l]{18}\makecell[r]{(1163.635)} &  \makecell[l]{0}\makecell[r]{(1162.670)} \\ 
106  &  1375228477454  & \makecell[l]{\textbf{100}}\makecell[r]{(\textbf{2.745})} &  \makecell[l]{\textbf{100}}\makecell[r]{(140.282)} &  \makecell[l]{\textbf{100}}\makecell[r]{(821.318)} &  \makecell[l]{97}\makecell[r]{(1034.791)} &  \makecell[l]{76}\makecell[r]{(1485.375)} &  \makecell[l]{0}\makecell[r]{(10.090)} \\ 
108  &  1581487595537  & \makecell[l]{\textbf{100}}\makecell[r]{(\textbf{5.361})} &  \makecell[l]{99}\makecell[r]{(450.271)} &  \makecell[l]{99}\makecell[r]{(657.236)} &  \makecell[l]{85}\makecell[r]{(1129.557)} &  \makecell[l]{95}\makecell[r]{(1050.945)} &  \makecell[l]{0}\makecell[r]{(32.390)} \\ 
109  &  1718976905230  & \makecell[l]{\textbf{100}}\makecell[r]{(\textbf{3.911})} &  \makecell[l]{1}\makecell[r]{(1306.312)} &  \makecell[l]{\textbf{100}}\makecell[r]{(551.843)} &  \makecell[l]{59}\makecell[r]{(1248.325)} &  \makecell[l]{88}\makecell[r]{(1310.915)} &  \makecell[l]{0}\makecell[r]{(3352.550)} \\ 
110  &  1512768094224  & \makecell[l]{\textbf{100}}\makecell[r]{(\textbf{3.801})} &  \makecell[l]{\textbf{100}}\makecell[r]{(391.202)} &  \makecell[l]{\textbf{100}}\makecell[r]{(15.364)} &  \makecell[l]{\textbf{100}}\makecell[r]{(13.357)} &  \makecell[l]{\textbf{100}}\makecell[r]{(17.000)} &  \makecell[l]{0}\makecell[r]{(2269.480)} \\ 
111  &  2544013377548  & \makecell[l]{\textbf{100}}\makecell[r]{(\textbf{57.323})} &  \makecell[l]{0}\makecell[r]{(1567.153)} &  \makecell[l]{1}\makecell[r]{(1567.423)} &  \makecell[l]{0}\makecell[r]{(1808.095)} &  \makecell[l]{0}\makecell[r]{(1519.593)} &  \makecell[l]{0}\makecell[r]{(98.890)} \\ 
112  &  2475277107217  & \makecell[l]{\textbf{100}}\makecell[r]{(\textbf{30.221})} &  \makecell[l]{0}\makecell[r]{(1749.538)} &  \makecell[l]{5}\makecell[r]{(1692.161)} &  \makecell[l]{0}\makecell[r]{(1550.411)} &  \makecell[l]{0}\makecell[r]{(1652.289)} &  \makecell[l]{0}\makecell[r]{(3593.620)} \\ 
113  &  2200298487823  & \makecell[l]{\textbf{100}}\makecell[r]{(\textbf{8.621})} &  \makecell[l]{0}\makecell[r]{(1556.683)} &  \makecell[l]{13}\makecell[r]{(1357.006)} &  \makecell[l]{16}\makecell[r]{(1540.398)} &  \makecell[l]{3}\makecell[r]{(1510.060)} &  \makecell[l]{\textbf{100}}\makecell[r]{(48.390)} \\ 
115  &  1581588209685  & \makecell[l]{\textbf{100}}\makecell[r]{(\textbf{15.338})} &  \makecell[l]{2}\makecell[r]{(1787.766)} &  \makecell[l]{30}\makecell[r]{(1404.084)} &  \makecell[l]{87}\makecell[r]{(1038.877)} &  \makecell[l]{33}\makecell[r]{(1581.393)} &  \makecell[l]{0}\makecell[r]{(3150.560)} \\ 
117  &  1925303123981  & \makecell[l]{\textbf{100}}\makecell[r]{(\textbf{7.908})} &  \makecell[l]{0}\makecell[r]{(1474.054)} &  \makecell[l]{97}\makecell[r]{(1214.428)} &  \makecell[l]{79}\makecell[r]{(1456.617)} &  \makecell[l]{64}\makecell[r]{(1177.945)} &  \makecell[l]{0}\makecell[r]{(974.580)} \\ 
118  &  2406641475604  & \makecell[l]{\textbf{100}}\makecell[r]{(\textbf{15.660})} &  \makecell[l]{0}\makecell[r]{(1670.186)} &  \makecell[l]{0}\makecell[r]{(1715.431)} &  \makecell[l]{0}\makecell[r]{(1591.963)} &  \makecell[l]{0}\makecell[r]{(1529.837)} &  \makecell[l]{\textbf{100}}\makecell[r]{(1443.190)} \\ 
120  &  2337821335572  & \makecell[l]{\textbf{100}}\makecell[r]{(\textbf{12.455})} &  \makecell[l]{0}\makecell[r]{(1433.344)} &  \makecell[l]{19}\makecell[r]{(1570.731)} &  \makecell[l]{4}\makecell[r]{(1409.346)} &  \makecell[l]{3}\makecell[r]{(1737.625)} &  \makecell[l]{0}\makecell[r]{(162.680)} \\ 
 \hline
\end{tabular}

\end{center}
\end{table*}

\subsubsection {Results on DIMACS Benchmark.}
Table \ref{tab:dimacs} shows the comparative results of PbO-MWC and its competitors on DIMACS benchmark. From table \ref{tab:dimacs}, PbO-MWC provides a performance advantage in terms of solution-quality and running time. \textbf{On training set}, PbO-MWC achieves the best $W_{avg}$ with shortest averaged time on 6 of 10 instances. \textbf{On all 11 test instances}, PbO-MWC gets the best ${W_{max}}$ and ${W_{avg}}$ for all of them. Notice that on `MANN\textunderscore a81', PbO-MWC found $w_{max}$=\textbf{111,355}. So far as we know, this is a new best-known solution. In terms of running time, PbO-MWC has obvious advantage on most of the instances except three instances where the complete solver has advantage. 

\begin{table*} [t]
\begin{center}
\caption{Experimental results on REF benchmark. For all instances, each solver was performed with its optimized configuration trained on 3 instances in the upper part.}\label{tab:ref}
\fontsize{\mytablefontsize}{\mytablebaselineskip\baselineskip}\selectfont\setlength{\tabcolsep}{\kestabcolsep}
\begin{tabular}{l  c c c c c c c}

\hline
 
\multirow{2}*{Graph}  & \multirow{2}*{{\em solBest}} & PbO-MWC & MN/TS & LSCC & LSCC+BMS & RRWL & TSM-MWC \\ 
	  &  & \multicolumn{1}{c}{{\em\#Suc}($t_{avg}$)} & \multicolumn{1}{c}{{\em\#Suc}($t_{avg}$)} & \multicolumn{1}{c}{{\em\#Suc}($t_{avg}$)} & \multicolumn{1}{c}{{\em\#Suc}($t_{avg}$)}& \multicolumn{1}{c}{{\em\#Suc}($t_{avg}$)} & 
	  \multicolumn{1}{c}{{\em\#Suc}({\em time})}\\

\hline\hline

ref-60-1000  &  743  & \makecell[l]{100}\makecell[r]{(5.978)} &  \makecell[l]{100}\makecell[r]{(\textbf{0.096})} &  \makecell[l]{100}\makecell[r]{(2.604)} &  \makecell[l]{100}\makecell[r]{(2.815)} &  \makecell[l]{100}\makecell[r]{(3.587)} &  \makecell[l]{100}\makecell[r]{(277.880)} \\ 
ref-60-230-0  &  506  & \makecell[l]{\textbf{100}}\makecell[r]{(\textbf{37.242})} &  \makecell[l]{87}\makecell[r]{(1045.383)} &  \makecell[l]{0}\makecell[r]{(770.114)} &  \makecell[l]{0}\makecell[r]{(758.782)} &  \makecell[l]{0}\makecell[r]{(1240.335)} &  \makecell[l]{0}\makecell[r]{(2477.980)} \\ 
ref-60-500-7  &  700  & \makecell[l]{\textbf{100}}\makecell[r]{(\textbf{5.511})} &  \makecell[l]{\textbf{100}}\makecell[r]{(58.170)} &  \makecell[l]{2}\makecell[r]{(57.438)} &  \makecell[l]{2}\makecell[r]{(76.478)} &  \makecell[l]{0}\makecell[r]{(39.422)} &  \makecell[l]{0}\makecell[r]{(399.550)} \\  

\hline

ref-60-10000  &  768  & \makecell[l]{100}\makecell[r]{(14.181)} &  \makecell[l]{100}\makecell[r]{(\textbf{0.032})} &  \makecell[l]{100}\makecell[r]{(0.097)} &  \makecell[l]{100}\makecell[r]{(0.118)} &  \makecell[l]{100}\makecell[r]{(0.527)} &  \makecell[l]{100}\makecell[r]{(12.890)} \\ 
ref-60-230-1  &  506  & \makecell[l]{\textbf{100}}\makecell[r]{(\textbf{14.083})} &  \makecell[l]{\textbf{100}}\makecell[r]{(52.042)} &  \makecell[l]{2}\makecell[r]{(259.120)} &  \makecell[l]{0}\makecell[r]{(201.255)} &  \makecell[l]{0}\makecell[r]{(992.196)} &  \makecell[l]{0}\makecell[r]{(830.790)} \\ 
ref-60-230-2  &  524  & \makecell[l]{\textbf{100}}\makecell[r]{(\textbf{0.030})} &  \makecell[l]{\textbf{100}}\makecell[r]{(0.249)} &  \makecell[l]{\textbf{100}}\makecell[r]{(337.901)} &  \makecell[l]{\textbf{100}}\makecell[r]{(260.969)} &  \makecell[l]{98}\makecell[r]{(629.556)} &  \makecell[l]{0}\makecell[r]{(3054.720)} \\ 
ref-60-230-3  &  502  & \makecell[l]{\textbf{100}}\makecell[r]{(\textbf{0.072})} &  \makecell[l]{\textbf{100}}\makecell[r]{(0.244)} &  \makecell[l]{\textbf{100}}\makecell[r]{(173.394)} &  \makecell[l]{\textbf{100}}\makecell[r]{(194.756)} &  \makecell[l]{97}\makecell[r]{(800.821)} &  \makecell[l]{0}\makecell[r]{(1781.070)} \\ 
ref-60-230-4  &  504  & \makecell[l]{\textbf{100}}\makecell[r]{(\textbf{0.098})} &  \makecell[l]{\textbf{100}}\makecell[r]{(0.712)} &  \makecell[l]{98}\makecell[r]{(855.686)} &  \makecell[l]{\textbf{100}}\makecell[r]{(499.303)} &  \makecell[l]{61}\makecell[r]{(1091.397)} &  \makecell[l]{0}\makecell[r]{(1511.560)} \\ 
ref-60-230-5  &  503  & \makecell[l]{\textbf{100}}\makecell[r]{(\textbf{0.280})} &  \makecell[l]{\textbf{100}}\makecell[r]{(0.339)} &  \makecell[l]{\textbf{100}}\makecell[r]{(362.522)} &  \makecell[l]{\textbf{100}}\makecell[r]{(429.003)} &  \makecell[l]{82}\makecell[r]{(1048.387)} &  \makecell[l]{0}\makecell[r]{(128.590)} \\ 
ref-60-230-6  &  505  & \makecell[l]{\textbf{100}}\makecell[r]{(\textbf{0.027})} &  \makecell[l]{\textbf{100}}\makecell[r]{(0.091)} &  \makecell[l]{\textbf{100}}\makecell[r]{(135.267)} &  \makecell[l]{\textbf{100}}\makecell[r]{(79.723)} &  \makecell[l]{\textbf{100}}\makecell[r]{(548.991)} &  \makecell[l]{0}\makecell[r]{(3351.150)} \\ 
ref-60-230-7  &  506  & \makecell[l]{\textbf{100}}\makecell[r]{(\textbf{2.353})} &  \makecell[l]{\textbf{100}}\makecell[r]{(6.780)} &  \makecell[l]{14}\makecell[r]{(432.156)} &  \makecell[l]{15}\makecell[r]{(429.282)} &  \makecell[l]{3}\makecell[r]{(936.927)} &  \makecell[l]{0}\makecell[r]{(2569.800)} \\ 
ref-60-230-8  &  494  & \makecell[l]{\textbf{100}}\makecell[r]{(\textbf{0.149})} &  \makecell[l]{\textbf{100}}\makecell[r]{(0.310)} &  \makecell[l]{\textbf{100}}\makecell[r]{(355.894)} &  \makecell[l]{\textbf{100}}\makecell[r]{(364.748)} &  \makecell[l]{93}\makecell[r]{(1022.198)} &  \makecell[l]{0}\makecell[r]{(1219.250)} \\ 
ref-60-230-9  &  526  & \makecell[l]{\textbf{100}}\makecell[r]{(\textbf{0.352})} &  \makecell[l]{\textbf{100}}\makecell[r]{(2.677)} &  \makecell[l]{98}\makecell[r]{(1065.225)} &  \makecell[l]{\textbf{100}}\makecell[r]{(599.256)} &  \makecell[l]{61}\makecell[r]{(1003.228)} &  \makecell[l]{0}\makecell[r]{(44.490)} \\ 
ref-60-300  &  599  & \makecell[l]{\textbf{100}}\makecell[r]{(\textbf{0.035})} &  \makecell[l]{\textbf{100}}\makecell[r]{(0.232)} &  \makecell[l]{\textbf{100}}\makecell[r]{(183.132)} &  \makecell[l]{\textbf{100}}\makecell[r]{(103.767)} &  \makecell[l]{99}\makecell[r]{(708.523)} &  \makecell[l]{0}\makecell[r]{(3567.480)} \\ 
ref-60-500-0  &  704  & \makecell[l]{\textbf{100}}\makecell[r]{(\textbf{9.023})} &  \makecell[l]{48}\makecell[r]{(735.908)} &  \makecell[l]{0}\makecell[r]{(353.410)} &  \makecell[l]{0}\makecell[r]{(121.051)} &  \makecell[l]{0}\makecell[r]{(325.049)} &  \makecell[l]{0}\makecell[r]{(3552.830)} \\ 
ref-60-500-1  &  709  & \makecell[l]{\textbf{100}}\makecell[r]{(\textbf{0.020})} &  \makecell[l]{\textbf{100}}\makecell[r]{(0.090)} &  \makecell[l]{\textbf{100}}\makecell[r]{(9.942)} &  \makecell[l]{\textbf{100}}\makecell[r]{(3.448)} &  \makecell[l]{\textbf{100}}\makecell[r]{(11.789)} &  \makecell[l]{0}\makecell[r]{(908.640)} \\ 
ref-60-500-2  &  702  & \makecell[l]{\textbf{100}}\makecell[r]{(\textbf{0.834})} &  \makecell[l]{\textbf{100}}\makecell[r]{(12.230)} &  \makecell[l]{4}\makecell[r]{(145.902)} &  \makecell[l]{3}\makecell[r]{(155.079)} &  \makecell[l]{1}\makecell[r]{(254.645)} &  \makecell[l]{0}\makecell[r]{(68.080)} \\ 
ref-60-500-4  &  690  & \makecell[l]{\textbf{100}}\makecell[r]{(\textbf{0.168})} &  \makecell[l]{\textbf{100}}\makecell[r]{(0.727)} &  \makecell[l]{\textbf{100}}\makecell[r]{(339.175)} &  \makecell[l]{\textbf{100}}\makecell[r]{(109.800)} &  \makecell[l]{\textbf{100}}\makecell[r]{(804.839)} &  \makecell[l]{0}\makecell[r]{(500.710)} \\ 
ref-60-500-6  &  715  & \makecell[l]{\textbf{100}}\makecell[r]{(\textbf{0.035})} &  \makecell[l]{\textbf{100}}\makecell[r]{(0.089)} &  \makecell[l]{\textbf{100}}\makecell[r]{(8.521)} &  \makecell[l]{\textbf{100}}\makecell[r]{(6.583)} &  \makecell[l]{\textbf{100}}\makecell[r]{(18.712)} &  \makecell[l]{0}\makecell[r]{(3161.160)} \\ 
ref-60-500-8  &  714  & \makecell[l]{\textbf{100}}\makecell[r]{(\textbf{0.021})} &  \makecell[l]{\textbf{100}}\makecell[r]{(0.100)} &  \makecell[l]{\textbf{100}}\makecell[r]{(286.912)} &  \makecell[l]{\textbf{100}}\makecell[r]{(123.516)} &  \makecell[l]{\textbf{100}}\makecell[r]{(850.038)} &  \makecell[l]{0}\makecell[r]{(411.690)} \\ 
ref-60-500-9  &  704  & \makecell[l]{\textbf{100}}\makecell[r]{(\textbf{0.923})} &  \makecell[l]{\textbf{100}}\makecell[r]{(5.372)} &  \makecell[l]{17}\makecell[r]{(366.494)} &  \makecell[l]{47}\makecell[r]{(905.199)} &  \makecell[l]{9}\makecell[r]{(204.615)} &  \makecell[l]{0}\makecell[r]{(642.830)} \\ 
ref-60-500  &  704  & \makecell[l]{\textbf{100}}\makecell[r]{(\textbf{0.476})} &  \makecell[l]{\textbf{100}}\makecell[r]{(0.940)} &  \makecell[l]{\textbf{100}}\makecell[r]{(141.320)} &  \makecell[l]{\textbf{100}}\makecell[r]{(81.310)} &  \makecell[l]{\textbf{100}}\makecell[r]{(59.510)} &  \makecell[l]{0}\makecell[r]{(2790.890)} \\ 
 
 \hline
\end{tabular}

\end{center}
\end{table*}

\begin{table*} [t]
\begin{center}
\caption{The {\em avgPAR10} on each benchmark.}\label{tab:sum}
\fontsize{\mytablefontsize}{\mytablebaselineskip\baselineskip}\selectfont\setlength{\tabcolsep}{\kestabcolsep}
\begin{tabular}{l  c c c c c c c}

\hline
 
\multirow{2}*{Benchmark}  & \multirow{2}*{$Num$} & PbO-MWC & MN/TS & LSCC & LSCC+BMS & RRWL & TSM-MWC \\ 
	  &  & \multicolumn{1}{c}{{\em avgPAR10}} & \multicolumn{1}{c}{{\em avgPAR10}} & \multicolumn{1}{c}{{\em avgPAR10}} & \multicolumn{1}{c}{{\em avgPAR10}}& \multicolumn{1}{c}{{\em avgPAR10}} & \multicolumn{1}{c}{{\em avgPAR10}}\\

\hline\hline

BHOSLIB & 40 & \makecell[c]{\textbf{228.682}}  & \makecell[c]{7234.972}  & \makecell[c]{18707.556}  &  \makecell[c]{19151.201}  &  \makecell[c]{18743.424}  &  \makecell[c]{32519.003}  \\
DIMACS & 80 & \makecell[c]{\textbf{903.982}}  & \makecell[c]{1379.709}  & \makecell[c]{1885.051}  &  \makecell[c]{1897.256}  &  \makecell[c]{1770.183}  &  \makecell[c]{5113.066}  \\
KES & 42 & \makecell[c]{\textbf{6.154}}  & \makecell[c]{16728.550}  & \makecell[c]{6728.632}  &  \makecell[c]{8526.514}  &  \makecell[c]{10163.870}  &  \makecell[c]{20723.301}  \\
REF & 29 & \makecell[c]{\textbf{3.170}}  & \makecell[c]{872.675}  & \makecell[c]{8423.623}  &  \makecell[c]{8000.062}  &  \makecell[c]{10168.337}  &  \makecell[c]{27320.371}  \\
 
 \hline
\end{tabular}

\end{center}
\end{table*}

\subsubsection {Results on Kidney Exchange Scheme (KES) Benchmark}
The comparative results of PbO-MWC and its competitors on KES benchmark are illustrated in Table \ref{tab:kes}. From Table \ref{tab:kes}, our PbO-MWC algorithm performs much better than its competitors. 
\textbf{On training set}, PbO-MWC is the only solver achieves a 100\% {\em success rate} with the shortest $t_{avg}$ on all 10 instances. \textbf{On all 27 test instances}, PbO-MWC reaches a 100\% {\em success rate} for all of them, while the figure is 10, 16, 12, 13 and 11 for MN/TS, LSCC, LSCC+BMS, RRWL and TSM-MWC, respectively. The total {\em\#Suc} of PbO-MWC is 2700, while the figure is 1273, 2117, 2010, 1884 and 1100 for its competitors respectively.
In terms of running time, on 21 of the 27 instances, PbO-MWC can achieve {\em solBest} with $t_{avg}$\textless 10 seconds. 
The averaged time of PbO-MWC is 7.625, while the figure is 810.750, 637.970, 748.197, 797.203 and 736.626 for its competitors
respectively.

\subsubsection {Results on Research Excellence Framework (REF) Benchmark}
The results shown in Table \ref{tab:ref} indicate that PbO-MWC outperforms its competitors on REF benchmark. 
\textbf{On training set}, PbO-MWC is the only solver achieves a 100\% {\em success rate} on all 3 instances. On `ref-60-23-0' instance, PbO-MWC reaches {\em\#Suc}=100 with $t_{avg}$=37.242, MN/TS, the second best solver on this instance, reaches {\em\#Suc}=87 with $t_{avg}$=1045.383. \textbf{On all 19 test instances}, PbO-MWC reaches a 100\% {\em success rate} for all of them, while the figure is 18, 12, 14, 7 and 1 for MN/TS, LSCC, LSCC+BMS, RRWL and TSM-MWC, respectively. On `ref-60-500-0' instance, PbO-MWC is the only solver that can achieve a 100\% {\em success rate}, and the $t_{avg}$ of PbO-MWC is much shorter than that of its competitors. 
In terms of running time, PbO-MWC achieves {\em solBest} with shortest $t_{avg}$ on 18 of 19 instances. The averaged time of PbO-MWC is 2.272, while the figure is 43.114, 305.898, 245.693, 595.366 and 1584.664 for its competitors
respectively.

\begin{figure*}[htbp]
\centering

\subfigure[PAR10 between PbO-MWC and PbO-MWC (Default) on BHOSLIB.]{
\begin{minipage}[t]{0.5\linewidth}
\centering
\includegraphics[width=2in]{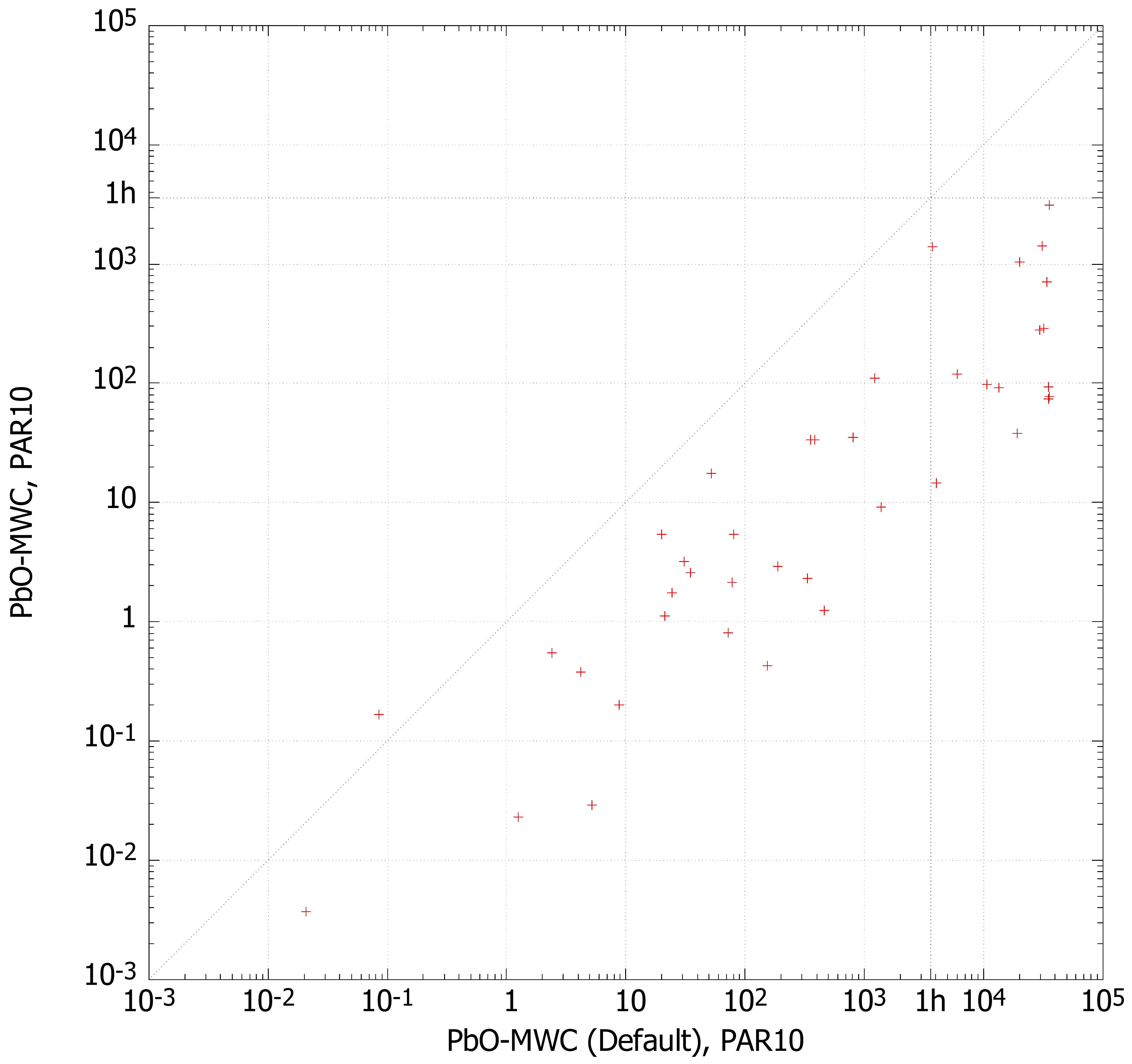}
\end{minipage}%
}%
\subfigure[PAR10 between PbO-MWC and PbO-MWC (Default) on DIMACS.]{
\begin{minipage}[t]{0.5\linewidth}
\centering
\includegraphics[width=2in]{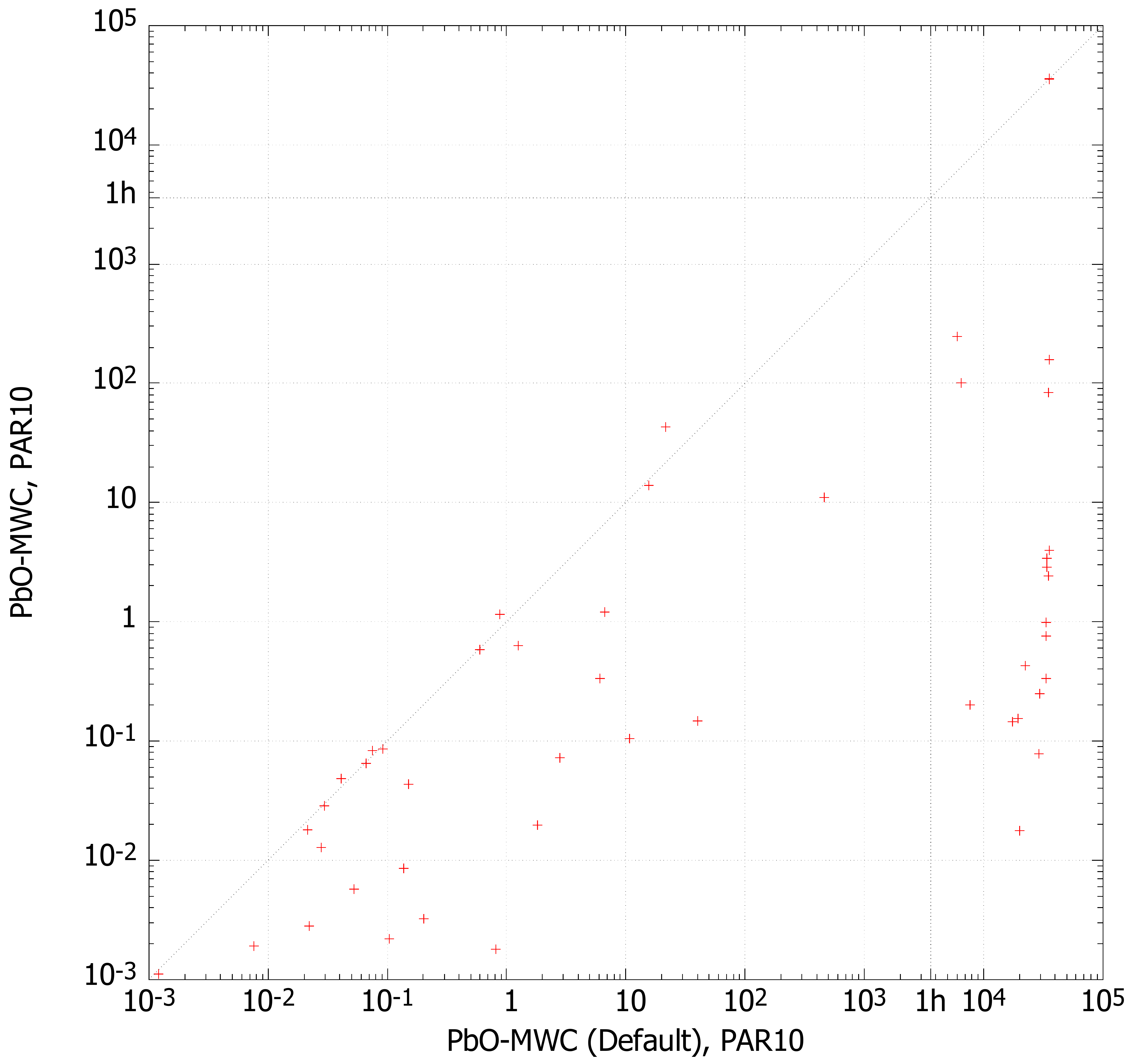}
\end{minipage}%
}%

\subfigure[PAR10 between PbO-MWC and PbO-MWC (Default) on KES.]{
\begin{minipage}[t]{0.5\linewidth}
\centering
\includegraphics[width=2in]{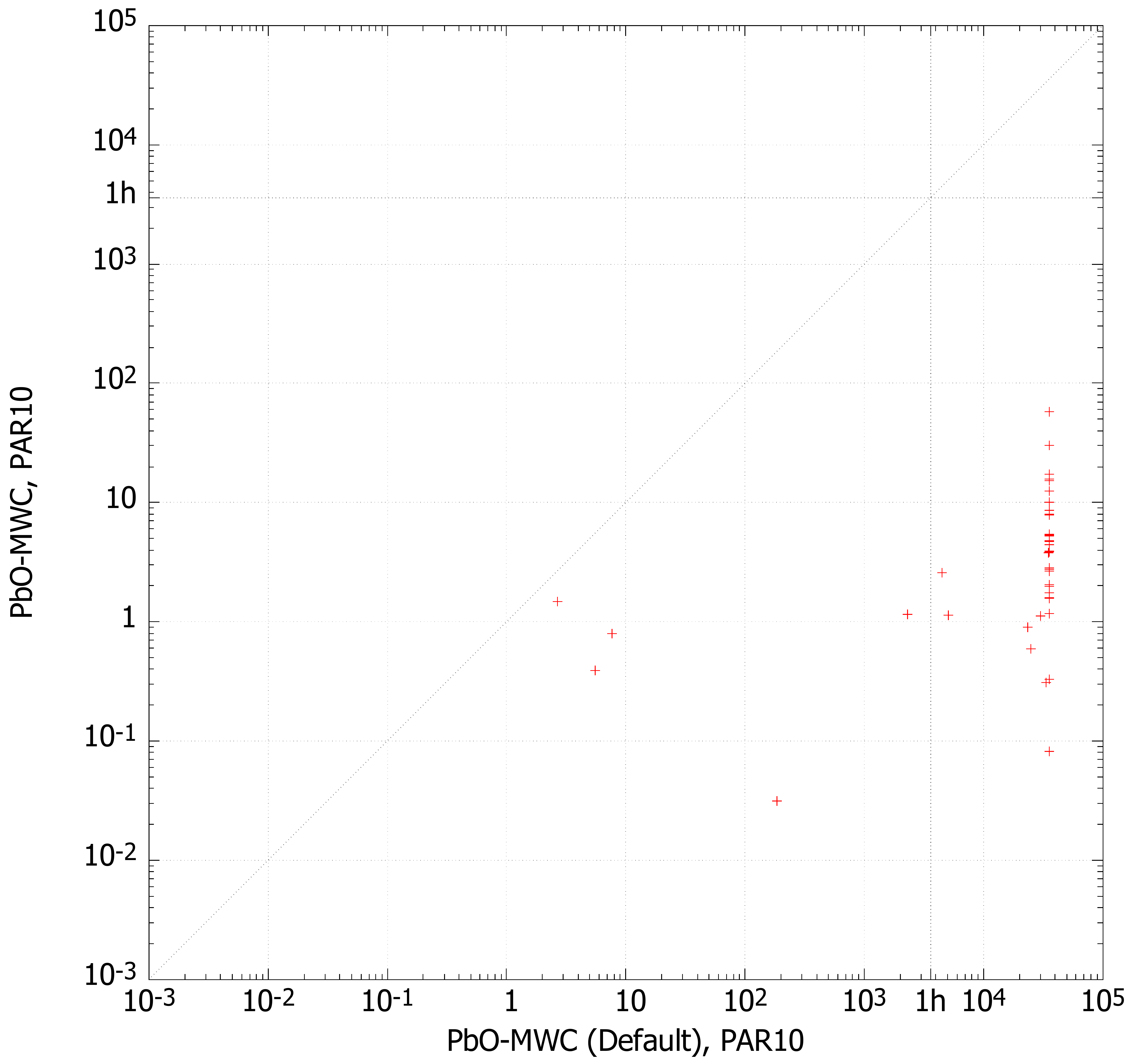}
\end{minipage}
}%
\subfigure[PAR10 between PbO-MWC and PbO-MWC (Default) on REF.]{
\begin{minipage}[t]{0.5\linewidth}
\centering
\includegraphics[width=2in]{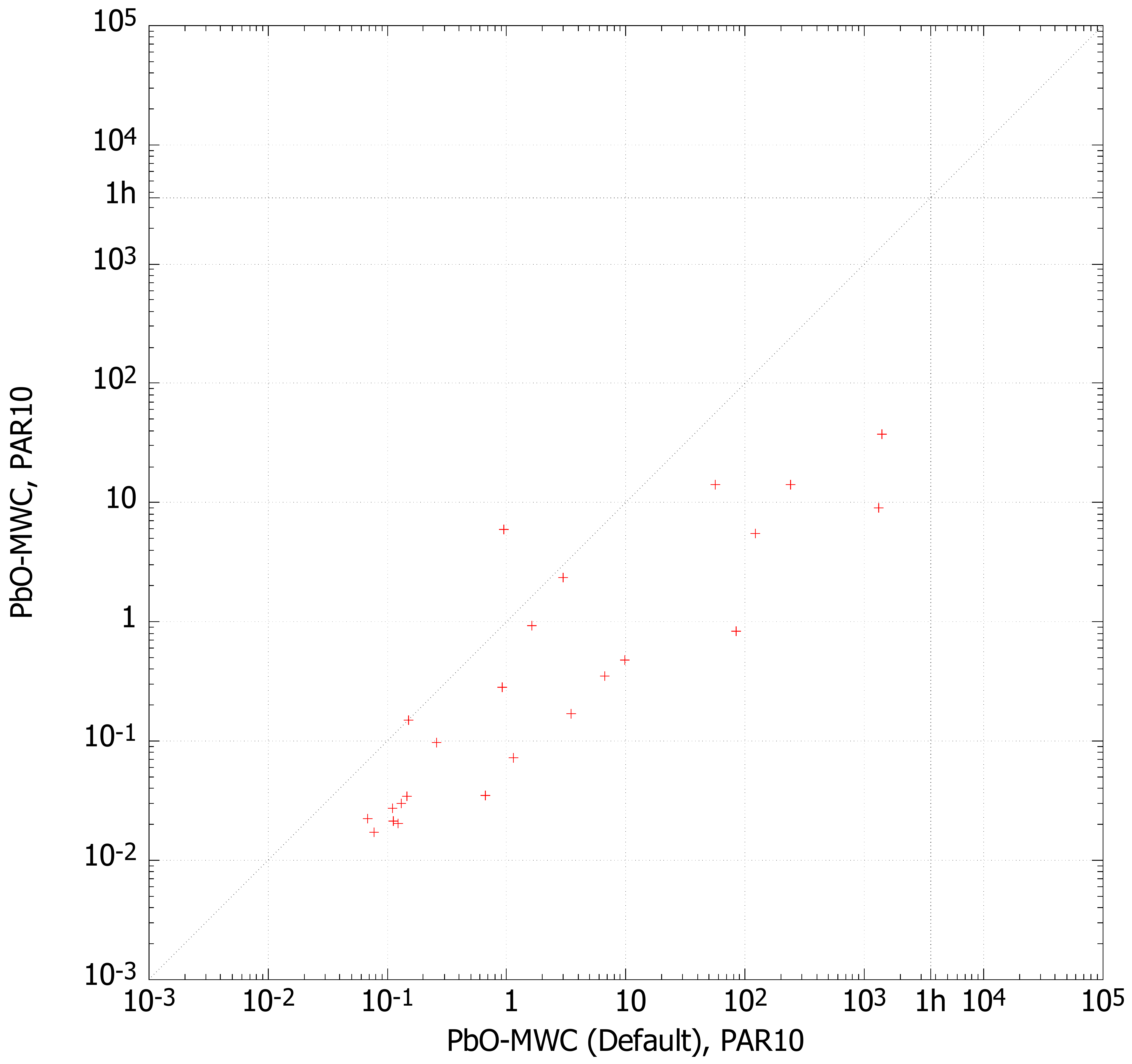}
\end{minipage}
}%
 
\centering
\caption{Scatter plots of PAR10 between PbO-MWC and PbO-MWC(Default) on four benchmarks. (1h [3600 seconds] is the cutoff time of each run.)}
\label{fig:default}
\end{figure*}

\subsubsection{An Overview of Results on All Benchmarks}
We summarize all the results in Table \ref{tab:sum}. Table \ref{tab:sum} shows that PbO-MWC outperforms all its competitors in terms of {\em avgPAR10} on all four benchmarks. On BHOSLIB, DIMACS, KES and REF, the ratio of {\em avgPAR10} of the best performing competitor to {\em avgPAR10} of PbO-MWC is 31.64, 1.53, 1093.38 and 275.29, respectively. The performance of state-of-the-art solver for MVWCP on BHOSLIB benchmark and two benchmarks that are transformed from real-world problems has been improved remarkably.

\subsubsection{The effect of automatically configuring PbO-MWC}
To illustrate the effect of automatic configuration of PbO-MWC, we report the performance comparison between PbO-MWC and PbO-MWC (Default) on the four benchmarks, as shown in Figure \ref{fig:default}. Figure \ref{fig:default} clearly illustrates that configuration leads to performance improvements on a large majority of instances on all the four benchmarks.

\subsubsection{The speculations on the effectiveness of different strategies}
Based on the extensive experiments, we made some speculations. We presumed that random walk strategy plays an important role on BHOSLIB. Among the three prohibition mechanisms utilized in PbO-MWC, SCC strategy performs best on DIMACS. 
In our experiments, all efficient configurations for KES contain BMS strategy with a parameter {\em bms\textunderscore num} that are less than 10. On REF, the optimized configuration includes TabuCC strategy, which works well on the REF benchmark. We consider an effective framework on REF benchmark has the following properties: restarting local search with a lower probability, including BMS strategy with a lower {\em bms\textunderscore num} and no random walk strategy is included.

\section{Conclusions and Future work}
\label{sec:conclusions}
In this work, we proposed a parametric SLS framework for MVWCP,  called PbO-MWC, which contains many effective techniques and restarts the local search process with a certain probability when it getting stuck local optima. We used the automated algorithm configuration procedure SMAC to configure PbO-MWC and its competitors. We conducted experiments to compare the performance of all the configured solvers on four benchmarks. On BHOSLIB, KES and REF, the ratio of {\em avgPAR10} of the second best solver to {\em avgPAR10} of PbO-MWC is 31.64, 1093.38 and 275.29 respectively. On `MANN\textunderscore a81' from DIMACS, a new optimal solution (\textbf{111,355}) is found by PbO-MWC.

In the future, we plan to do some research to understand which strategies are best suit for particular problem types, and design novel strategies to integrate them into PbO-MWC framework to further improve the performance.
In addition, we would like to conduct experiments on more benchmarks that are transformed from real-world problems.




\bibliographystyle{elsarticle-num}
\bibliography{pbomwc}

\begin{thebibliography}{10}
\expandafter\ifx\csname url\endcsname\relax
  \def\url#1{\texttt{#1}}\fi
\expandafter\ifx\csname urlprefix\endcsname\relax\def\urlprefix{URL }\fi
\expandafter\ifx\csname href\endcsname\relax
  \def\href#1#2{#2} \def\path#1{#1}\fi

\bibitem{Karp1972Reducibility}
R.~M. Karp, Reducibility among combinatorial problems, Journal of Symbolic
  Logic 40~(4) (1972) 618--619.

\bibitem{Ballard1982Computer}
D.~H. Ballard, C.~M. Brown, Computer vision (1982) 145.

\bibitem{Park1996An}
K.~Park, K.~Lee, S.~Park, An extended formulation approach to the edge-weighted
  maximal clique problem, European Journal of Operational Research 95~(3)
  (1996) 671--682.

\bibitem{Balasundaram2006Graph}
B.~Balasundaram, S.~Butenko, Graph Domination, Coloring and Cliques in
  Telecommunications, 2006.

\bibitem{ostergaard2001new}
P.~R. {\"O}sterg{\aa}rd, A new algorithm for the maximum-weight clique problem,
  Nordic Journal of Computing 8~(4) (2001) 424--436.

\bibitem{yamaguchi2008new}
K.~Yamaguchi, S.~Masuda, A new exact algorithm for the maximum weight clique
  problem, in: ITC-CSCC: International Technical Conference on Circuits
  Systems, Computers and Communications, 2008, pp. 317--320.

\bibitem{fang2016exact}
Z.~Fang, C.-M. Li, K.~Xu, An exact algorithm based on maxsat reasoning for the
  maximum weight clique problem, Journal of Artificial Intelligence Research 55
  (2016) 799--833.

\bibitem{jiang2017exact}
H.~Jiang, C.-M. Li, F.~Many{\`{a}}, An exact algorithm for the maximum weight
  clique problem in large graphs., in: AAAI, 2017, pp. 830--838.

\bibitem{jiang2018two}
H.~Jiang, C.-M. Li, Y.~Liu, F.~Many{\`{a}}, A two-stage maxsat reasoning
  approach for the maximum weight clique problem, in: Thirty-Second AAAI
  Conference on Artificial Intelligence, 2018.

\bibitem{battiti2001reactive}
R.~Battiti, M.~Protasi, Reactive local search for the maximum clique problem 1,
  Algorithmica 29~(4) (2001) 610--637.

\bibitem{grosso2004combining}
A.~Grosso, M.~Locatelli, F.~Della~Croce, Combining swaps and node weights in an
  adaptive greedy approach for the maximum clique problem, Journal of
  Heuristics 10~(2) (2004) 135--152.

\bibitem{pullan2006dynamic}
W.~Pullan, H.~H. Hoos, Dynamic local search for the maximum clique problem,
  Journal of Artificial Intelligence Research 25 (2006) 159--185.

\bibitem{pullan2008approximating}
W.~Pullan, Approximating the maximum vertex/edge weighted clique using local
  search, Journal of Heuristics 14~(2) (2008) 117--134.

\bibitem{pullan2011cooperating}
W.~Pullan, F.~Mascia, M.~Brunato, Cooperating local search for the maximum
  clique problem, Journal of Heuristics 17~(2) (2011) 181--199.

\bibitem{wu2012multi}
Q.~Wu, J.-K. Hao, F.~Glover, Multi-neighborhood tabu search for the maximum
  weight clique problem, Annals of Operations Research 196~(1) (2012) 611--634.

\bibitem{WangCY2016}
Y.~Wang, S.~Cai, M.~Yin, Two efficient local search algorithms for maximum
  weight clique problem, in: Proceedings of AAAI 2016, 2016, pp. 805--811.

\bibitem{CaiL2016}
S.~Cai, J.~Lin, Fast solving maximum weight clique problem in massive graphs,
  in: Proceedings of IJCAI 2016, 2016, pp. 568--574.

\bibitem{zhou2017push}
Y.~Zhou, J.-K. Hao, A.~Go{\"e}ffon, Push: A generalized operator for the
  maximum vertex weight clique problem, European Journal of Operational
  Research 257~(1) (2017) 41--54.

\bibitem{fan2017restart}
Y.~Fan, N.~Li, C.~Li, Z.~Ma, L.~J. Latecki, K.~Su, Restart and random walk in
  local search for maximum vertex weight cliques with evaluations in clustering
  aggregation, in: Proc. of International Joint Conference on Artificial
  Intelligence (IJCAI), 2017, pp. 622--630.

\bibitem{XuEtAl05}
K.~Xu, F.~Boussemart, F.~Hemery, C.~Lecoutre, A simple model to generate hard
  satisfiable instances, in: Proceedings of {IJCAI} 2005, 2005, pp. 337--342.

\bibitem{XuEtAl07}
K.~Xu, F.~Boussemart, F.~Hemery, C.~Lecoutre, Random constraint satisfaction:
  Easy generation of hard (satisfiable) instances, Artificial Intelligence
  171~(8-9) (2007) 514--534.

\bibitem{dimacs26}
D.~S. Johnson, M.~A. Trick (Eds.), Cliques, Coloring, and Satisfiability,
  Vol.~26 of {DIMACS} Series in Discrete Mathematics and Theoretical Computer
  Science, {DIMACS/AMS}, 1996.

\bibitem{hoos2012programming}
H.~H. Hoos, Programming by optimization., Commun. ACM 55~(2) (2012) 70--80.

\bibitem{KhuEtAl16}
A.~R. KhudaBukhsh, L.~Xu, H.~H. Hoos, K.~Leyton{-}Brown, {SATenstein}:
  Automatically building local search {SAT} solvers from components, Artificial
  Intelligence 232 (2016) 20--42.

\bibitem{hutter2010automated}
F.~Hutter, H.~H. Hoos, K.~Leyton-Brown, Automated configuration of mixed
  integer programming solvers, in: International Conference on Integration of
  Artificial Intelligence (AI) and Operations Research (OR) Techniques in
  Constraint Programming, Springer, 2010, pp. 186--202.

\bibitem{MVC2019luo}
C.~Luo, H.~H.Hoos, S.~Cai, Q.~Lin, H.~Zhang, D.~Zhang, Local search with
  efficient automatic configuration for minimum vertex cover, in: Proceedings
  of IJCAI 2019, 2019.

\bibitem{ansotegui2009gender}
C.~Ans{\'o}tegui, M.~Sellmann, K.~Tierney, A gender-based genetic algorithm for
  the automatic configuration of algorithms, in: International Conference on
  Principles and Practice of Constraint Programming, Springer, 2009, pp.
  142--157.

\bibitem{hutter2009paramils}
F.~Hutter, H.~H. Hoos, K.~Leyton-Brown, T.~St{\"u}tzle, {ParamILS}: an
  automatic algorithm configuration framework, Journal of Artificial
  Intelligence Research 36~(1) (2009) 267--306.

\bibitem{HutterHL2011}
F.~Hutter, H.~H. Hoos, K.~Leyton{-}Brown, Sequential model-based optimization
  for general algorithm configuration, in: Proceedings of LION 2011, 2011, pp.
  507--523.

\bibitem{hutter2017configurable}
F.~Hutter, M.~Lindauer, A.~Balint, S.~Bayless, H.~Hoos, K.~Leyton-Brown, The
  configurable sat solver challenge (cssc), Artificial Intelligence 243 (2017)
  1--25.

\bibitem{xu2011hydra}
L.~Xu, F.~Hutter, H.~H. Hoos, K.~Leyton-Brown, Hydra-mip: Automated algorithm
  configuration and selection for mixed integer programming, in: RCRA workshop
  on experimental evaluation of algorithms for solving problems with
  combinatorial explosion at the international joint conference on artificial
  intelligence (IJCAI), 2011, pp. 16--30.

\bibitem{vallati2013automatic}
M.~Vallati, C.~Fawcett, A.~E. Gerevini, H.~Hoos, A.~Saetti, Automatic
  generation of efficient domain-optimized planners from generic parametrized
  planners, in: Sixth Annual Symposium on Combinatorial Search, 2013.

\bibitem{birattari2010f}
M.~Birattari, Z.~Yuan, P.~Balaprakash, T.~St{\"u}tzle, F-race and iterated
  f-race: An overview, in: Experimental methods for the analysis of
  optimization algorithms, Springer, 2010, pp. 311--336.

\bibitem{L2016The}
M.~López-Ibáñez, J.~Dubois-Lacoste, L.~P. Cáceres, M.~Birattari,
  T.~Stützle, The irace package: Iterated racing for automatic algorithm
  configuration, Operations Research Perspectives 3 (2016) 43--58.

\bibitem{HoosS2004}
H.~H. Hoos, T.~St{\"{u}}tzle, Stochastic Local Search: Foundations {\&}
  Applications, Elsevier / Morgan Kaufmann, 2004.

\bibitem{glover1989tabupart1}
F.~Glover, Tabu search?part i, ORSA Journal on computing 1~(3) (1989) 190--206.

\bibitem{caiSS2011MVC}
S.~Cai, K.~Su, A.~Sattar, Local search with edge weighting and configuration
  checking heuristics for minimum vertex cover, Artificial Intelligence
  175~(9-10) (2011) 1672--1696.

\bibitem{Cai2015}
S.~Cai, Balance between complexity and quality: Local search for minimum vertex
  cover in massive graphs, in: Proceedings of IJCAI 2015, 2015, pp. 747--753.

\bibitem{johnson1996cliques}
D.~S. Johnson, M.~A. Trick, Cliques, coloring, and satisfiability: second
  DIMACS implementation challenge, October 11-13, 1993, Vol.~26, American
  Mathematical Soc., 1996.

\bibitem{richter2007stochastic}
S.~Richter, M.~Helmert, C.~Gretton, A stochastic local search approach to
  vertex cover, in: Annual Conference on Artificial Intelligence, Springer,
  2007, pp. 412--426.

\bibitem{li2017minimization}
C.-M. Li, H.~Jiang, F.~Many{\`a}, On minimization of the number of branches in
  branch-and-bound algorithms for the maximum clique problem, Computers \&
  Operations Research 84 (2017) 1--15.

\bibitem{mccreesh2017maximum}
C.~McCreesh, P.~Prosser, K.~Simpson, J.~Trimble, On maximum weight clique
  algorithms, and how they are evaluated, in: International Conference on
  Principles and Practice of Constraint Programming, Springer, 2017, pp.
  206--225.

\end{thebibliography}


%
%

\end{document}